\pdfoutput=1
\documentclass{article}
\PassOptionsToPackage{numbers,sort&compress}{natbib}
\usepackage[preprint]{neurips_2026}
\usepackage[utf8]{inputenc}
\usepackage[T1]{fontenc}
\usepackage{amsmath}
\usepackage{graphicx}
\usepackage{booktabs}
\usepackage{tabularx}
\usepackage{xcolor}
\usepackage{microtype}
\usepackage{enumitem}
\usepackage[font=small,labelfont=bf]{caption}
\usepackage[hidelinks]{hyperref}
\usepackage{url}
\usepackage{longtable}
\usepackage{float}
\usepackage{amssymb}
\setlist{itemsep=0.15em,topsep=0.3em}

\newcommand{\NEvents}{426}
\newcommand{\NCatalog}{65}

\newcommand{\NHarvest}{139,135}
\newcommand{\NPool}{6,939}

\newcommand{\NGeneral}{1,598}
\newcommand{\NAdopted}{37}
\newcommand{\NIndependent}{18}
\newcommand{\NOnlyUnknown}{7}
\newcommand{\NThree}{7}
\newcommand{\NDefault}{1}
\newcommand{\NExceptions}{8}
\newcommand{\NGeneralPlus}{1,606}
\newcommand{\NAdoptedPlus}{45}
\newcommand{\NThreePlus}{9}
\newcommand{\NEarly}{769}
\newcommand{\NEarlyAdopted}{31}

\newcommand{\NLate}{829}
\newcommand{\NLateAdopted}{6}
\newcommand{\NCoreComp}{187}
\newcommand{\NCoreCompAdopted}{10}
\newcommand{\NCoreCompThree}{3}
\newcommand{\NCrowdComp}{554}
\newcommand{\NCrowdCompAdopted}{4}

\newcommand{\LagMedian}{30}
\newcommand{\NLags}{49}
\newcommand{\NLagsUpper}{4}
\newcommand{\NRecentIncluded}{54}
\newcommand{\NLagsNoDate}{1}
\newcommand{\NLagNegative}{5}
\newcommand{\PaperLagMedian}{30.5}
\newcommand{\NPaperLags}{42}
\newcommand{\NSingle}{38}
\newcommand{\NSingleSelf}{13}

\newcommand{\NWideLandings}{70}
\newcommand{\NWideAuthor}{6}
\newcommand{\NWideNotAuthor}{40}
\newcommand{\NWideUnknown}{18}
\newcommand{\NWideNoPaper}{6}
\newcommand{\GSPOtrl}{3}

\newcommand{\NJulyOthers}{43}
\newcommand{\NJulyOthersAdopted}{2}

\newcommand{\NMethodMonthlyMin}{144}
\newcommand{\CISPOLag}{135}
\newcommand{\NCatalogInWindow}{46}
\newcommand{\NCatalogInPool}{45}
\newcommand{\CCCases}{31}

\newcommand{\CCControls}{150}

\newcommand{\CCCasesAdded}{8}

\newcommand{\CCNoVOne}{1}

\newcommand{\CCBigThirtyCase}{12}
\newcommand{\CCBigThirtyCaseN}{29}
\newcommand{\CCBigThirtyCtrl}{14}
\newcommand{\CCBigThirtyCtrlN}{141}
\newcommand{\CCBigThirtyCaseWorst}{12}
\newcommand{\CCBigThirtyCtrlWorst}{23}

\newcommand{\CCMoeCase}{3}

\newcommand{\CCMoeCtrl}{1}

\newcommand{\CCCoefCase}{9}
\newcommand{\CCCoefCaseN}{31}
\newcommand{\CCCoefCtrl}{13}
\newcommand{\CCCoefCtrlN}{149}
\newcommand{\CCCoefCaseWorst}{9}
\newcommand{\CCCoefCtrlWorst}{14}

\newcommand{\CCBigFourteenCase}{13}
\newcommand{\CCBigFourteenCaseN}{29}
\newcommand{\CCBigFourteenCtrl}{19}
\newcommand{\CCBigFourteenCtrlN}{141}

\newcommand{\CCNoRepBigThirtyCase}{12}
\newcommand{\CCNoRepBigThirtyCaseN}{29}
\newcommand{\CCNoRepBigThirtyCtrl}{13}
\newcommand{\CCNoRepBigThirtyCtrlN}{139}
\newcommand{\CCIndBigThirtyCase}{10}
\newcommand{\CCIndBigThirtyCaseN}{25}
\newcommand{\CCIndBigThirtyCtrl}{11}
\newcommand{\CCIndBigThirtyCtrlN}{90}
\newcommand{\CCAllBigThirtyCase}{17}
\newcommand{\CCAllBigThirtyCaseN}{35}
\newcommand{\CCOrigBigThirtyCase}{11}
\newcommand{\CCOrigBigThirtyCaseN}{22}
\newcommand{\CCNoRepCoefCase}{9}
\newcommand{\CCNoRepCoefCaseN}{31}
\newcommand{\CCNoRepCoefCtrl}{13}
\newcommand{\CCNoRepCoefCtrlN}{147}
\newcommand{\CCIndCoefCase}{7}
\newcommand{\CCIndCoefCaseN}{26}
\newcommand{\CCIndCoefCtrl}{9}
\newcommand{\CCIndCoefCtrlN}{93}
\newcommand{\CCAllCoefCase}{9}
\newcommand{\CCAllCoefCaseN}{38}
\newcommand{\CCOrigCoefCase}{8}
\newcommand{\CCOrigCoefCaseN}{23}

\newcommand{\CCRecodedCtrls}{16}
\newcommand{\CCRecBigThirtyCtrl}{14}
\newcommand{\CCRecBigThirtyCtrlN}{141}
\newcommand{\CCRecCoefCtrl}{12}
\newcommand{\CCRecCoefCtrlN}{149}
\newcommand{\CCBigThirtyCasePct}{41}
\newcommand{\CCBigThirtyCtrlPct}{10}
\newcommand{\CCCoefCasePct}{29}
\newcommand{\CCCoefCtrlPct}{9}
\newcommand{\CCPipelineCasePct}{26}
\newcommand{\CCPipelineCtrlPct}{53}
\newcommand{\CCCodeCasePct}{44}
\newcommand{\CCCodeCtrlPct}{44}
\newcommand{\CCIndustryAnyCasePct}{90}
\newcommand{\CCIndustryAnyCtrlPct}{64}
\newcommand{\CCFailureCasePct}{84}
\newcommand{\CCFailureCtrlPct}{70}
\newcommand{\CCComputeCasePct}{19}
\newcommand{\CCComputeCtrlPct}{29}
\newcommand{\CCBroadCasePct}{55}
\newcommand{\CCBroadCtrlPct}{48}

\newcommand{\CCOwnCase}{5}
\newcommand{\CCNoOrgBigThirtyCase}{6}
\newcommand{\CCNoOrgBigThirtyCaseN}{19}
\newcommand{\CCNoOrgBigThirtyCtrl}{9}
\newcommand{\CCNoOrgBigThirtyCtrlN}{119}
\newcommand{\CCNoOwnBigThirtyCase}{9}
\newcommand{\CCNoOwnBigThirtyCaseN}{24}
\newcommand{\CCNoOrgCoefCase}{6}
\newcommand{\CCNoOrgCoefCaseN}{21}
\newcommand{\CCNoOrgCoefCtrl}{13}
\newcommand{\CCNoOrgCoefCtrlN}{128}
\newcommand{\CCNoOwnCoefCase}{8}
\newcommand{\CCNoOwnCoefCaseN}{26}

\newcommand{\CCOrgCasePct}{32}
\newcommand{\CCOrgCtrlPct}{15}

\newcommand{\CCQOneCase}{7}

\newcommand{\CCQOneCtrl}{32}
\newcommand{\CCQTwoCase}{10}

\newcommand{\CCQTwoCtrl}{54}
\newcommand{\CCQThreeCase}{7}

\newcommand{\CCQThreeCtrl}{27}
\newcommand{\CCQFourCase}{7}

\newcommand{\CCQFourCtrl}{37}
\newcommand{\CCPMainBigThirty}{$<$0.001}
\newcommand{\CCPAllBigThirty}{$<$0.001}
\newcommand{\CCPNoRepBigThirty}{$<$0.001}
\newcommand{\CCPIndBigThirty}{0.003}
\newcommand{\CCPWorstBigThirty}{0.005}
\newcommand{\CCPOrigBigThirty}{$<$0.001}
\newcommand{\CCPNoOwnBigThirty}{0.001}
\newcommand{\CCPNoOrgBigThirty}{0.007}
\newcommand{\CCPFourteenBigThirty}{$<$0.001}
\newcommand{\CCPMainCoef}{0.004}
\newcommand{\CCPAllCoef}{0.020}
\newcommand{\CCPNoRepCoef}{0.005}
\newcommand{\CCPIndCoef}{0.045}
\newcommand{\CCPWorstCoef}{0.006}
\newcommand{\CCPOrigCoef}{0.002}
\newcommand{\CCPNoOwnCoef}{0.004}
\newcommand{\CCPNoOrgCoef}{0.031}
\newcommand{\NOrgIncluded}{11}
\newcommand{\NOwnIncluded}{6}

\newcommand{\NAuthorOnly}{12}
\newcommand{\VerlRecipeCommit}{\texttt{81d782765e}}
\newcommand{\VerlRecipeDate}{2026-10-13}
\newcommand{\MinStars}{2,000}

\newcommand{\NFwExtraNew}{2}

\newcommand{\RateWithExtra}{2.4}
\newcommand{\NReportsRead}{70}
\newcommand{\NReportsMethod}{30}
\newcommand{\NReportsOther}{40}
\newcommand{\NReportsMethodGen}{25}
\newcommand{\NReportsOtherGen}{6}

\newcommand{\NReportAdjWidePct}{13}
\newcommand{\RateWide}{2.1}
\newcommand{\RateMain}{2.3}

\newcommand{\NFollowEligible}{1,155}
\newcommand{\NFollowIncluded}{30}
\newcommand{\LandFirst}{3}
\newcommand{\LandLast}{9}
\newcommand{\LandPeak}{10}

\newcommand{\QRateFirst}{6.9}
\newcommand{\QRateLast}{1.3}

\newcommand{\QSupplyFirst}{72}
\newcommand{\QSupplyLast}{386}
\newcommand{\QSupplyRatio}{5.4}
\newcommand{\QIncMin}{5}
\newcommand{\QIncMax}{7}

\newcommand{\ReviewGenHit}{68}
\newcommand{\ReviewGenN}{76}
\newcommand{\ReviewEmbHit}{3}
\newcommand{\ReviewEmbN}{16}
\newcommand{\ReviewOthHit}{1}
\newcommand{\ReviewOthN}{108}
\newcommand{\ReviewEstimate}{1,590}

\newcommand{\PrverlLand}{27}
\newcommand{\PrverlMaint}{1}
\newcommand{\PrverlAtt}{20}
\newcommand{\PrverlAttOpen}{12}

\newcommand{\PrverlAttStale}{7}

\newcommand{\PrTRLLand}{23}
\newcommand{\PrTRLMaint}{13}
\newcommand{\PrTRLAtt}{18}

\newcommand{\PrTRLAttDeclined}{10}

\newcommand{\PrslimeLand}{7}
\newcommand{\PrslimeMaint}{6}

\newcommand{\PrOpenRLHFLand}{5}
\newcommand{\PrOpenRLHFMaint}{4}

\newcommand{\NRecentLandings}{114}
\newcommand{\NRemoved}{4}

\newcommand{\NThreads}{26,960}
\newcommand{\NNotIncluded}{1,561}
\newcommand{\NNoRecord}{1,500}
\newcommand{\NNoRecordPct}{96}

\newcommand{\NNameSearch}{726}
\newcommand{\NNameCand}{23}
\newcommand{\NNameRefers}{13}
\newcommand{\NNameCited}{5}
\newcommand{\NNameProposal}{8}

\newcommand{\NNoRecordAfterPct}{95}
\newcommand{\NNoRecordExtrapPct}{94}
\newcommand{\NImplPapers}{25}
\newcommand{\NRequestPapers}{15}
\newcommand{\NCitedOnly}{21}
\newcommand{\NAttemptThreads}{48}
\newcommand{\NAttemptPapers}{40}
\newcommand{\NRefusedPapers}{12}
\newcommand{\NAttemptPRs}{28}
\newcommand{\NAttemptRequests}{20}
\newcommand{\NPROpen}{17}
\newcommand{\NPROpenNoReply}{13}

\newcommand{\NPRDeclined}{4}
\newcommand{\NPRRedirected}{2}

\newcommand{\NReqStale}{8}
\newcommand{\NReqDeclined}{4}
\newcommand{\NReqRedirected}{2}

\newcommand{\NNoMergerReply}{17}

\newcommand{\NNoMergerReplyNone}{16}
\newcommand{\OpenPRMedianAge}{196}

\newcommand{\NLandingEvents}{94}
\newcommand{\NLandingPRs}{88}
\newcommand{\NLandingPriorRequest}{22}
\newcommand{\NLandingAuthor}{21}
\newcommand{\NPRPriorMerger}{29}

\newcommand{\PRDaysOpen}{6.5}

\newcommand{\PRCoreLines}{117}

\newcommand{\AttemptCoreLines}{162}
\newcommand{\CoreLinesCoef}{88}

\newcommand{\CoreLinesTerm}{80}

\newcommand{\CoreLinesPipe}{141}

\newcommand{\NAuditCited}{51}
\newcommand{\NAuditBefore}{28}
\newcommand{\NAuditAdded}{9}
\newcommand{\NAuditList}{10}
\newcommand{\NAuditUnmatched}{23}
\newcommand{\NAuditReport}{2}
\newcommand{\NAuditOther}{2}
\newcommand{\AgreeThreads}{48}
\newcommand{\AgreeOutcome}{40}
\newcommand{\RevMN}{88}

\newcommand{\RevMSilent}{17}
\newcommand{\RevMAny}{60}
\newcommand{\RevMOpen}{41}
\newcommand{\RevMDS}{6}

\newcommand{\RevMResults}{26}

\newcommand{\RevMPrior}{57}
\newcommand{\RevMFixed}{49}
\newcommand{\RevUN}{28}

\newcommand{\RevUSilent}{14}

\newcommand{\RevUDS}{5}

\newcommand{\RevUResults}{9}

\newcommand{\RevRN}{20}

\newcommand{\RevRDS}{6}

\newcommand{\RevOtherAny}{18}
\newcommand{\RevOtherPrior}{17}

\newcommand{\RevNeMoMN}{13}
\newcommand{\RevNeMoMReq}{8}

\newcommand{\NIssueUniverse}{7,075}
\newcommand{\NIssueScreened}{648}
\newcommand{\NIssueFailure}{48}
\newcommand{\NIssueRequest}{59}

\newcommand{\NIssueRecall}{250}

\newcommand{\NIssueTrainInferMismatch}{12}

\newcommand{\NIssueInstability}{23}

\newcommand{\NReqDynamicSampling}{8}
\newcommand{\NReqRoutingReplay}{4}
\newcommand{\NReqMismatch}{6}

\newcommand{\NSupplyTrainInferMismatch}{20}

\newcommand{\NCatalogTrainInferMismatch}{8}

\newcommand{\NSupplySparseReward}{218}

\newcommand{\NSupplyDiversityLoss}{156}

\newcommand{\NReports}{13}
\newcommand{\NReportsUseAny}{6}
\newcommand{\NReportsNamedFramework}{4}

\newcommand{\NReportsSource}{2}
\newcommand{\NReportsVariant}{3}
\newcommand{\NReportsNone}{4}
\newcommand{\NReportsNoKL}{4}
\newcommand{\ReportFirst}{DeepSeek-R1}
\newcommand{\ReportLast}{Ling and Ring 2.6}
\newcommand{\NReportsMethodsUsed}{8}
\newcommand{\NWideMethods}{13}

\newcommand{\NRepDapoTokenLoss}{3}

\newcommand{\RecallHit}{30}
\newcommand{\RecallAll}{32}
\newcommand{\RecallCleanHit}{29}
\newcommand{\RecallCleanAll}{30}
\newcommand{\RepeatAgree}{129}
\newcommand{\RepeatAll}{140}
\newcommand{\ReviewAgree}{188}
\newcommand{\ReviewAll}{200}

\newcommand{\WideDefaultsNote}{Of these, truncated importance sampling (TIS) is a conditional default in TRL and token-level loss aggregation is a default in TRL.}

\makeatletter
\renewcommand{\@noticestring}{Technical report, version 1, 20 October 2026. Data and code: \url{https://github.com/pth2002/rl-method-tracker}, archived as \href{https://doi.org/10.5281/zenodo.23495743}{doi:10.5281/zenodo.23495743}.}
\makeatother

\title{From RL Papers to Framework Code:\\ Inclusion and Review in Seven Open-Source Training Frameworks}
\author{Taiheng Pan\\
The University of Melbourne\\
\texttt{taiheng.pan@student.unimelb.edu.au}}

\begin{document}
\maketitle

\begin{abstract}
Practitioners use reinforcement learning methods for language models through training frameworks, so a method that a framework includes is one configuration change away for its users. We measure how often this happens. We trace seven open-source frameworks for RL post-training through their commit histories and link them to \NPool{} screened arXiv papers from January 2025 to June 2026. Of \NGeneral{} papers that propose a general RL training method, \NAdopted{} reached at least one framework. Across five quarterly cohorts of papers, the number included within 180 days stayed between \QIncMin{} and \QIncMax{} while the cohorts grew from \QSupplyFirst{} to \QSupplyLast{} papers. Compared with papers from the same quarters, included papers more often trained policies of 30B parameters or more, at \CCBigThirtyCasePct{} against \CCBigThirtyCtrlPct{} percent, and more often changed a coefficient, clip, mask or weight in the existing loss, at \CCCoefCasePct{} against \CCCoefCtrlPct{} percent. Both gaps hold without papers from the companies that maintain the frameworks, while code release was equally common in both groups. The frameworks' issues and pull requests show that few methods are turned away in public. More than nine in ten of the general methods no framework included have no thread there that cites or names them, and many proposals that do appear stay open without a reply. Two further records, the frameworks' issues and the technical reports of \NReports{} large open-weight models, show that users rarely report training failures in public and that the reports state \NReportsMethodsUsed{} of the \NWideMethods{} widely included methods. We release the data as a tracker that links training failures to framework implementations and lists each framework's contribution rules and earlier proposals.
\end{abstract}

\section{Introduction}

GSPO \citep{arxiv250718071} reached TRL \GSPOtrl{} days after its first arXiv version, and verl, OpenRLHF, ROLL and slime followed within a week. All seven frameworks in this study include it today. In the same month, \NJulyOthers{} other papers proposed a general RL training method for language models, and only \NJulyOthersAdopted{} of them reached any of the seven frameworks.

Research on RL post-training has grown fast: in each month of the first half of 2026, at least \NMethodMonthlyMin{} papers in our pool proposed an RL training method. Practitioners meet these methods through the frameworks they train with, and a framework decides whether a method is one configuration change away or a separate project. Publishing a method, releasing its code and having a framework include it are three separate outcomes, and the last one has not been measured.

We measure framework inclusion. A method counts as included when its implementation reaches the main branch of a tracked framework, and we record the tier of each inclusion: a default, a selectable algorithm, a switch, experimental code or an example recipe. Inclusion describes the code that frameworks ship, and Section~\ref{sec:limits} sets out its limits.

Inclusion is rare: only \NAdopted{} of \NGeneral{} papers that propose a general method reached one of seven frameworks (\RateMain{} percent). The included methods differ from the rest in two ways (Section~\ref{sec:what}). Their papers more often trained policies of 30B parameters or more, at \CCBigThirtyCasePct{} against \CCBigThirtyCtrlPct{} percent, and more often changed a coefficient, clip, mask or weight in the existing loss, at \CCCoefCasePct{} against \CCCoefCtrlPct{} percent. Code release was equally common, at \CCCodeCasePct{} percent of included papers and \CCCodeCtrlPct{} percent of others. The two gaps persist without model reports, within papers that have an industry author, and without papers from the companies that maintain the frameworks. They remain associations, and Section~\ref{sec:what} sets out their limits.

\begin{figure}[!t]
\centering
\includegraphics[width=\linewidth]{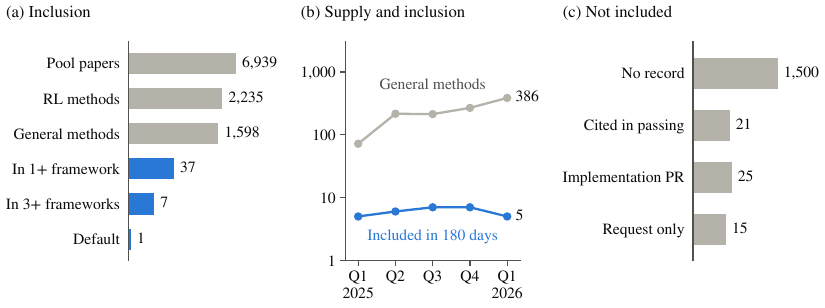}
\caption{\textbf{Inclusion, supply against intake, and the public record of the rest.} Unit: paper. (a) Papers from January 2025 to June 2026: the screened pool, papers that propose an RL training method, general methods, and general methods by the strongest inclusion they reached, on a log scale. (b) General methods by the quarter of their first arXiv version in grey, and those a framework included within 180 days of that version in blue, for the quarters with 180 days of exposure before the snapshot for every paper, on a log scale. (c) General methods that no framework included, grouped by the threads in the seven frameworks that link them, give their identifier or quote their title, on a log scale. These groups are categories and not stages.}
\label{fig:key}
\end{figure}

The intake did not keep pace with the supply. Across five quarterly cohorts, the number of general methods a framework included within 180 days stayed between \QIncMin{} and \QIncMax{}, while the cohorts grew from \QSupplyFirst{} to \QSupplyLast{} papers, a \QSupplyRatio{}-fold rise (Figure~\ref{fig:key}b). Counted by the date of the landing instead, first landings of general methods rose from \LandFirst{} a quarter at the start of 2025 to \LandPeak{} at its end. The frameworks took in more methods as the supply grew, but a shrinking share of each cohort.

Few of the other methods were turned away in public. We collected the issues, pull requests and discussions of all seven frameworks and searched them for every general method in the pool. Maintainers declined or redirected proposals for only \NRefusedPapers{} of the \NNotIncluded{} general methods that no framework included. Most never came up in a framework's issues or pull requests at all: \NNoRecordPct{} percent have no thread that cites them (Figure~\ref{fig:key}c), and a search by method name lowers this only to \NNoRecordAfterPct{} percent. Many proposals that do appear stay open without a reply. Merged pull requests also drew concerns, mostly about design and tests, and their authors often revised the code in response. Concerns about demand and maintenance burden appear mostly in proposals that were not merged, and most of those come from one framework (Section~\ref{sec:routes}).

Two further records place inclusion next to users and to large training runs. In the frameworks' issues and discussions, \NIssueFailure{} of the \NIssueUniverse{} threads opened since January 2025 report a training failure, and the failures they name are distributed differently from those that paper abstracts address (Section~\ref{sec:demand}). Technical reports of \NReports{} open-weight models with 100B parameters or more state \NReportsMethodsUsed{} of the \NWideMethods{} widely included methods, describe variants of others under their own names, and are the source of \NReportsSource{} (Section~\ref{sec:use}).

We release the data and code as a tracker. Its index links training failures, such as entropy collapse or a mismatch between rollout and training probabilities, to the implementations and settings in each framework. A second page gives each framework's contribution rules, the concerns its reviewers raised and the outcome of earlier proposals, as a starting point for authors who want a framework to include their method.

\section{Related work}

Studies of industry and academia in AI mostly work at the level of papers and topics. They track the growing share of industry in AI research \citep{ahmed2023industry} and the association between code availability and citation counts \citep{arxiv230810008}. Forks of the repositories that accompany ML papers rarely change the code, and fewer send changes back \citep{arxiv220311365}. These measures capture attention among researchers but do not observe whether a method reaches the code that trains models.

A second line asks whether reported gains hold up. Re-evaluations in recommendation \citep{arxiv190706902} and metric learning \citep{arxiv200308505} found that many improvements vanished under careful comparison, and broader critiques point to benchmark selection \citep{arxiv210707002} and scholarship practice \citep{arxiv180703341}. Implementation details matter as much in policy-gradient methods: code-level choices explain much of the reported gap between PPO and TRPO \citep{arxiv200512729}, and an implementation guide lists 37 such details for PPO alone \citep{huang2022details}. For RL with language models, controlled studies compare tricks under one implementation \citep{arxiv250808221} and test methods at large compute \citep{arxiv251013786}. These studies ask whether a method works. We ask whether frameworks include it, and we read the answer from their code.

A third line studies the decisions of open-source projects on outside contributions. Only a few factors drive the decision to merge a pull request, and few rejections are technical \citep{gousios2014pull}. Project managers weigh technical signals together with their social ties to the submitter \citep{tsay2014influence}, and integrators report trouble keeping quality high and ranking the contributions they receive \citep{gousios2015integrator}. Reviewers also look at signals beyond the code \citep{ford2019beyond}, and ties across an ecosystem shift acceptance \citep{meijer2025ecosystem}. These studies explain decisions about code contributions in general. We follow a narrower object: a method from a paper, traced from its evidence to its implementation in a training framework and to the public record of each proposal.

\section{Measuring inclusion}
\label{sec:inclusion}

\paragraph{Frameworks.} We study verl \citep{arxiv240919256} with its recipe repository, TRL \citep{vonwerra2020trl}, OpenRLHF \citep{arxiv240511143}, NeMo-RL \citep{nemorl2025}, ROLL \citep{arxiv250606122}, slime \citep{slime2025} and AReaL \citep{arxiv250524298}. All seven meet four criteria: they are built for RL post-training of language models, run their own training loop with several selectable RL algorithms, are developed in public, and have at least \MinStars{} GitHub stars. Four other frameworks meet the same criteria: open-instruct, SkyRL, prime-rl and RLinf (Table~\ref{tab:candidates}). At their latest commits they implement \NFwExtraNew{} general methods that none of the seven include, so adding them would raise the inclusion rate from \RateMain{} to \RateWithExtra{} percent. The snapshots of six frameworks date from 13 to 17 October 2026. OpenRLHF had no main-branch commit after 17 September 2026, so its snapshot is older (Table~\ref{tab:candidates}).

\paragraph{Algorithm locations.} Each framework exposes its RL algorithms in a few places. verl registers advantage estimators and policy losses by name, while TRL selects GRPO losses through a loss-type field and keeps newer trainers in an experimental folder. AReaL ships one example configuration per algorithm, and the other frameworks use command-line choices and configuration switches. We list these places for each framework, parse them at every commit on the main branch, and check the result against each framework's algorithm documentation at the snapshot.

\paragraph{Tiers.} We record the tier of each inclusion. A \emph{default} is on in a run that keeps every setting. A \emph{conditional default} is on by default once a related feature is enabled, such as generation with vLLM. A \emph{core} option is a selectable algorithm in the main code path, and a \emph{switch} modifies the main algorithm. \emph{Experimental} code and example \emph{recipes} sit outside the main path.

\paragraph{From options to methods.} Frameworks name the same method differently: clip-higher \citep{arxiv250314476} appears as \texttt{clip\_ratio\_high} in verl, \texttt{epsilon\_high} in TRL, \texttt{eps\_clip\_low\_high} in OpenRLHF and \texttt{eps-clip-high} in slime. We map every option name to one entry in a catalog of \NCatalog{} methods and check each source against its arXiv record. The parser reports every option it detects without a mapping, and the build fails until the option is mapped. An event is the addition or removal of an option, or a change of its default, on a main branch, and the history of the seven frameworks yields \NEvents{} events. Implementations outside the parsed places are the target of the documentation check and of the citation audit below.

\paragraph{Citation audit.} Parsing misses an implementation that sits under a name the catalog does not map, in a place we do not parse, or in a change to existing code. We therefore searched the full main-branch history of each framework for the arXiv identifiers of pool papers and kept every file that cites one. Framework files cite \NAuditCited{} general methods, including all \NAuditBefore{} that parsing had found. For \NAuditAdded{} of the other \NAuditUnmatched{}, a citation sits next to an implementation, and we added them to the catalog. Two of them add no option: a fix to the KL gradient in NeMo-RL and the trajectory-level agent pipeline of ROLL. We record such implementations from their commits, so the implementation stays the test of inclusion. The remaining papers appear in lists of projects built on a framework, in reference lists of technical reports, or as background (Appendix~\ref{app:record}).

\paragraph{Dates.} A date is the commit date on the main branch, an approximation of the day a change became available there. An option first seen in a repository's root commit, in a bulk code release, or in a commit that renamed an existing option has an unknown start, so we treat such a date as an upper bound. A \emph{landing} is the first commit that brings a method into one framework, and the \emph{first landing} of a method is its earliest landing in any framework. A later exact date from another framework never replaces an earlier upper bound. Lags run from the first arXiv version of the source, or from the post date for blog sources.

\paragraph{People.} For each landing we compare the git author of the introducing commit with the author list of the source. The comparison has four outcomes: a paper author, a named person outside the list, a handle we cannot resolve, and a source without an author list. When the commit names a pull request, we also record who opened it, who merged it and who reviewed it. We call a person a prior merger of a framework once they have merged a pull request there, comparing UTC timestamps with the event in question. Prior mergers stand in for maintainers, but the proxy misses reviewers without merge rights and people before their first merge, and it keeps people who have since left. We also match the affiliations on each paper's first page against the company that maintains each framework, such as ByteDance for verl and NVIDIA for NeMo-RL. OpenRLHF has no maintaining company.

\section{Measuring the supply of methods}
\label{sec:supply}

\paragraph{Screening.} Keyword search through the arXiv API misses papers that typeset method names in LaTeX. The DAPO abstract, for example, writes its name through bold initials. We therefore harvest the complete computer science metadata through the arXiv OAI-PMH interface and keep the \NHarvest{} records in cs.LG, cs.CL or cs.AI with identifiers from 2501 to 2606. A paper enters the pool when its title or abstract mentions language models and reinforcement learning, including the abbreviations RL and RFT. The pool holds \NPool{} papers and contains \NCatalogInPool{} of the \NCatalogInWindow{} catalog sources from this window. This check covers included sources only, so the pool is a screened set and not a complete census of RL papers.

\paragraph{Labels.} Each paper receives one of seven labels from its title and abstract under fixed rules (Appendix~\ref{app:rules}). The label of interest marks a new or modified training-time RL component. We call such a method \emph{general} when the component is the paper's main contribution and is presented for RL with language models at large. We call it \emph{embedded} when it serves a paper about one task, domain or application. The denominator is the set of general methods.

\paragraph{Checks.} Three checks address recognition, stability and the general rule. For recognition, the included sources were mixed into the first batches without marking, and \RecallCleanHit{} of the \RecallCleanAll{} sources not used as examples in the labelling rules were labelled as methods. The miss is a technical report that describes its method only in the body. For stability, \RepeatAll{} papers were labelled twice in separate batches and kept the same primary label \RepeatAgree{} times. For the general rule, a stratified sample of \ReviewAll{} papers was reviewed manually. Review and labels agree on \ReviewAgree{} of them for the split between general methods and all other papers, and weighting by the three label strata puts the number of general methods at about \ReviewEstimate{} (Appendix~\ref{app:review}).

\paragraph{Model reports.} Technical reports for new models often describe their RL changes only in the body, so we read \NReportsRead{} randomly drawn reports in full. The sample holds \NReportsMethod{} reports labelled as methods and \NReportsOther{} labelled otherwise, and it leaves out reports with included methods. A transferable RL component appears in the body of \NReportsMethodGen{} reports of the first group and \NReportsOtherGen{} of the second. The general rule asks for more than a transferable component, so these counts bound the effect of abstract-only labels from above. Counting every report with such a component as a general method would raise the denominator by at most about \NReportAdjWidePct{} percent and lower the inclusion rate from \RateMain{} to \RateWide{} percent.

\section{Inclusion rate}
\label{sec:how-much}

Few general methods reach a framework. Figure~\ref{fig:key}a follows the pool from all RL papers to framework defaults: \NAdopted{} of \NGeneral{} general methods reached at least one framework, a rate of \RateMain{} percent. Three or more frameworks include \NThree{} of them today, and \NDefault{} is an unconditional default.

Many methods were brought in by people outside the author list. For \NIndependent{} included papers, at least one landing came from a named commit author outside the paper's author list. Another \NAuthorOnly{} reached frameworks only through commits by their authors, and the remaining \NOnlyUnknown{} only through unresolved handles or sources without an author list.

Another \NExceptions{} included papers fall outside the general rule from their abstracts. Most are model reports or system papers with an included component in the body (Appendix~\ref{app:exceptions}). Adding them gives \NAdoptedPlus{} included papers out of \NGeneralPlus{}, with \NThreePlus{} in three or more frameworks.

Exposure time matters: papers from 2025 had at least nine months before the snapshot, and \NEarlyAdopted{} of \NEarly{} reached a framework. Papers from the first half of 2026 had less time, and \NLateAdopted{} of \NLate{} did. With a common window, \NFollowIncluded{} of the \NFollowEligible{} general methods with at least 180 days of exposure were included within 180 days. Counted by the quarter of the landing instead, first landings of general methods rose from \LandFirst{} in the first quarter of 2025 to \LandPeak{} in the fourth, and \LandLast{} landed in the first quarter of 2026. The intake grew while the share of each cohort included within 180 days fell from \QRateFirst{} to \QRateLast{} percent.

\begin{figure}[t]
\centering
\includegraphics[width=\linewidth]{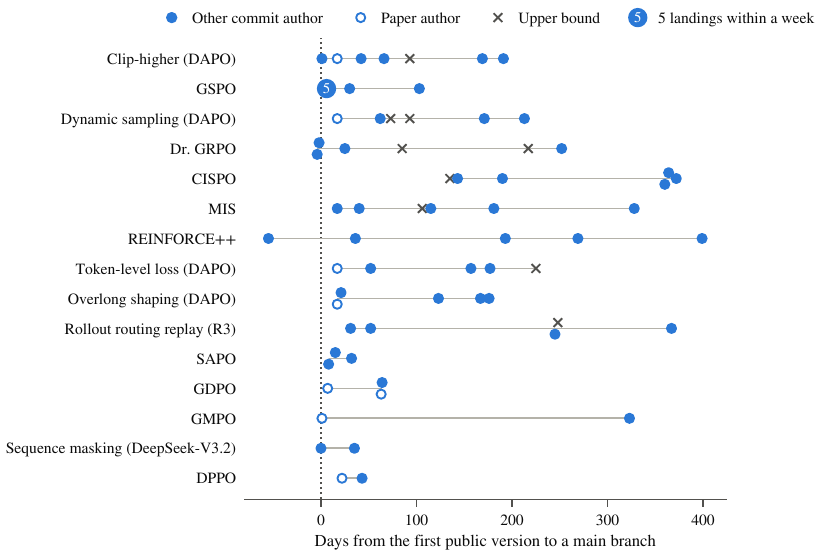}
\caption{\textbf{Time from first public version to a main branch.} Unit: catalog method with a dated source from 2025 or later that reached two or more frameworks. Figure~\ref{fig:landing-all} in the appendix shows every method. Each mark is one landing. Hollow dots mark commits by a paper author. Crosses mark landings in a bulk release or a renaming commit, and their dates are upper bounds. Landings within a few days of each other are spread vertically within their row. The larger mark with a number counts landings too close to draw apart: five frameworks took in GSPO between 3 and 7 days after its first version.}
\label{fig:landing}
\end{figure}

Inclusion is fast when it happens. Among the \NRecentIncluded{} catalog methods with a source from 2025 or later, \NLags{} have an exact first landing, and their median time from the first arXiv version to a main branch is \LagMedian{} days (Figure~\ref{fig:landing}). Another \NLagsUpper{} have only an upper bound, and \NLagsNoDate{} blog source is dated only by month. Counted once per source paper, the median is \PaperLagMedian{} days over \NPaperLags{} papers, and in \NLagNegative{} cases the commit predates the arXiv version. These medians describe included methods only. Among methods in three or more frameworks, CISPO \citep{arxiv250613585} is the slowest: it reached its first framework at most \CISPOLag{} days after its source, and TRL added it later in a commit named after a large comparison study \citep{arxiv251013786}.

Authorship changes with breadth. Among the \NSingle{} recent methods found in exactly one framework, \NSingleSelf{} first landed through a commit by a paper author, most of them as example recipes. Methods in three or more frameworks have \NWideLandings{} landings, and \NWideNotAuthor{} of these came from named people outside the author list. Paper authors made \NWideAuthor{}, unresolved handles \NWideUnknown{}, and the remaining \NWideNoPaper{} belong to blog sources without an author list.

Defaults are rare. TRL turned on token-level loss aggregation in August 2025 and sets the KL coefficient to zero by default. It also enables truncated importance sampling by default once generation with vLLM is switched on, and applies the importance-weighted KL correction of DeepSeek-V3.2 \citep{arxiv251202556} once the KL coefficient is non-zero. NeMo-RL applies the corrected KL gradient of \citet{arxiv250609477} whenever its KL penalty is non-zero (Section~\ref{sec:routes}). Table~\ref{tab:wide} lists the recent methods included in three or more frameworks.

\begin{table}[t]
\centering
\small
\caption{\textbf{Recent methods included in three or more frameworks.} Unit: catalog method. Days run from the first public version of the source to the first landing on a main branch. A $\le$ sign marks an upper bound, and negative values mean the commit predates the arXiv version. \WideDefaultsNote}
\label{tab:wide}
\begin{tabularx}{\linewidth}{>{\raggedright\arraybackslash}p{3.7cm} l l >{\raggedright\arraybackslash}p{1.7cm} r >{\raggedright\arraybackslash}X}
\toprule
Method & Source & First public & Frameworks & Days & Failure it addresses \\
\midrule
Clip-higher & \citep{arxiv250314476} & 2025-03-18 & 7 & 1 & entropy collapse \\
GSPO & \citep{arxiv250718071} & 2025-07-24 & 7 & 3 & MoE instability, extreme ratios \\
Dynamic sampling & \citep{arxiv250314476} & 2025-03-18 & 6 & 17 & groups without signal \\
Dr. GRPO & \citep{arxiv250320783} & 2025-03-26 & 6 & -4 & length bias \\
CISPO & \citep{arxiv250613585} & 2025-06-16 & 6 & $\le$135 & entropy collapse \\
Truncated importance sampling (TIS) & \citep{yao2025tis} & 2025-08 & 6 & -- & rollout-training mismatch \\
Masked importance sampling (MIS) & \citep{arxiv251018855} & 2025-10-21 & 6 & 17 & rollout-training mismatch \\
REINFORCE++ & \citep{arxiv250103262} & 2025-01-04 & 5 & -55 & value-model cost, gradient variance \\
Token-level loss aggregation & \citep{arxiv250314476} & 2025-03-18 & 5 & 17 & length bias \\
Overlong filtering and penalty & \citep{arxiv250314476} & 2025-03-18 & 5 & 17 & truncation noise \\
Rollout routing replay & \citep{arxiv251011370} & 2025-10-13 & 5 & 31 & MoE instability, rollout-training mismatch \\
SAPO & \citep{arxiv251120347} & 2025-11-25 & 3 & 8 & extreme ratios \\
GDPO & \citep{arxiv260105242} & 2026-01-08 & 3 & 7 & multi-reward collapse \\
\bottomrule
\end{tabularx}

\end{table}

\section{Features of included methods}
\label{sec:what}

\begin{figure}[!htbp]
\centering
\includegraphics[width=\linewidth]{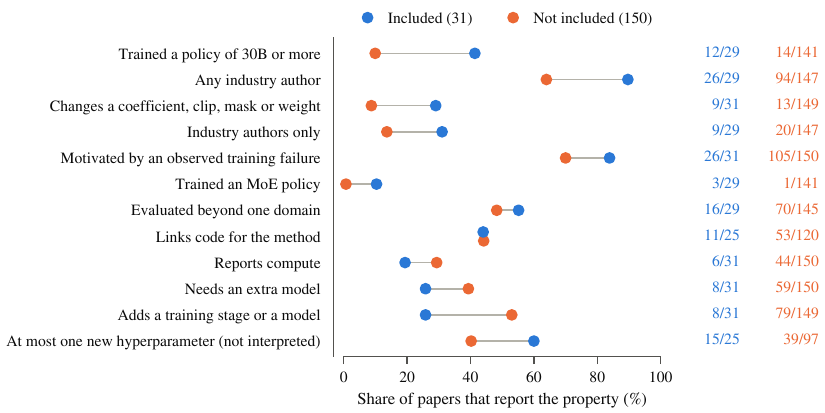}
\caption{\textbf{Coded fields, included against non-included general methods from 2025.} Unit: paper. Each row compares the share among included papers with the share among papers from the same quarters that no framework included. The right columns give the counts, included first. Two shares within three points of each other are drawn one above the other. Papers that do not report a property are left out of that row.}
\label{fig:casecontrol-all}
\end{figure}

Figure~\ref{fig:casecontrol-all} compares the \CCCases{} included general methods from 2025 with \CCControls{} general methods from the same year that no framework included. The controls were drawn at random within each quarter, and fields were coded from the first arXiv version of each paper without access to inclusion status (Appendix~\ref{app:fulltext}). The citation audit found \CCCasesAdded{} of the included papers later, and a second reader recoded these papers blind. The comparison uses the blind codes, and Table~\ref{tab:sensitivity} gives the gaps without them. The comparison is descriptive, because scale, resources, team composition and breadth of experiments can vary together.

The figure shows every coded field. We single out two of them because they have the largest ratio between the groups among the fields that at least five papers show in each group. Mixture-of-experts policies differ by a larger ratio, but they appear in only \CCMoeCase{} included papers and \CCMoeCtrl{} other.

Scale shows the largest gap. Policies of 30B parameters or more appear in \CCBigThirtyCasePct{} percent of included papers and \CCBigThirtyCtrlPct{} percent of the others. The gap persists after excluding model reports, at \CCNoRepBigThirtyCase{} of \CCNoRepBigThirtyCaseN{} against \CCNoRepBigThirtyCtrl{} of \CCNoRepBigThirtyCtrlN{}, and within papers that have an industry author, at \CCIndBigThirtyCase{} of \CCIndBigThirtyCaseN{} against \CCIndBigThirtyCtrl{} of \CCIndBigThirtyCtrlN{}. Counting every unreported size against the gap still leaves \CCBigThirtyCaseWorst{} of \CCCases{} against \CCBigThirtyCtrlWorst{} of \CCControls{}.

The form of the change also differs. A coefficient, clip, mask or weight in the existing loss was the change in \CCCoefCasePct{} percent of included papers and \CCCoefCtrlPct{} percent of the others, while a new training stage or model was the change in \CCPipelineCasePct{} against \CCPipelineCtrlPct{} percent. The frameworks expose the widely included methods in the same compact form: one or two configuration fields or loss names, such as \texttt{clip\_ratio\_high} in verl and \texttt{epsilon\_high} in TRL for clip-higher (Table~\ref{tab:configs}). The compact form describes the method and not the size of its code, because pull requests for coefficient changes were about as large as those for new loss terms (Section~\ref{sec:routes}). The number of new hyperparameters is missing for many papers, and under the least favourable treatment of missing values its gap reverses, so we do not interpret it.

The two gaps hold outside the frameworks' own companies. Included papers come from a company that maintains one of the frameworks more often than others, at \CCOrgCasePct{} against \CCOrgCtrlPct{} percent, but only \CCOwnCase{} were included by that company's framework. Across all \NAdopted{} included papers, \NOrgIncluded{} come from such a company and \NOwnIncluded{} were included by its own framework, such as DAPO in verl and DLER in NeMo-RL. Inclusion also crosses company lines through the people who commit: all \NIndependent{} papers with an outside committer had such a landing in a framework outside their own company. Without any such paper in either group, large policies appear in \CCNoOrgBigThirtyCase{} of \CCNoOrgBigThirtyCaseN{} included papers against \CCNoOrgBigThirtyCtrl{} of \CCNoOrgBigThirtyCtrlN{} others, and coefficient changes in \CCNoOrgCoefCase{} of \CCNoOrgCoefCaseN{} against \CCNoOrgCoefCtrl{} of \CCNoOrgCoefCtrlN{} (Table~\ref{tab:sensitivity}).

Code release was equally common in both groups. The first version links training code in \CCCodeCasePct{} percent of included papers and \CCCodeCtrlPct{} percent of others, so a public implementation alone does not set the included methods apart. Other fields point the same way as the main gaps with smaller differences. A named training failure motivated \CCFailureCasePct{} percent of included papers and \CCFailureCtrlPct{} percent of others, and evaluation beyond one domain appears in \CCBroadCasePct{} against \CCBroadCtrlPct{} percent of them. Industry authors appear on \CCIndustryAnyCasePct{} percent of included papers and \CCIndustryAnyCtrlPct{} percent of others, while included papers reported compute less often, at \CCComputeCasePct{} against \CCComputeCtrlPct{} percent.

\begin{figure}[!htbp]
\centering
\includegraphics[width=\linewidth]{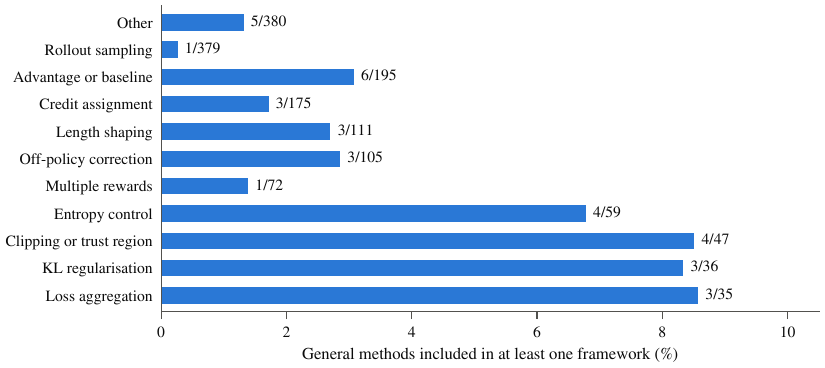}
\caption{\textbf{Inclusion by component.} Unit: general method, each with one primary component label. Share of general methods in each component that reached at least one framework, with counts. Rows are ordered by the number of general methods, and components with fewer than ten are omitted.}
\label{fig:components}
\end{figure}

Inclusion also differs by component (Figure~\ref{fig:components}). Methods that change clipping or the trust region, off-policy correction, or loss aggregation make up \NCoreComp{} of the general methods, and \NCoreCompAdopted{} of these reached a framework, \NCoreCompThree{} of them three or more. Rollout sampling is the most crowded component. Together with credit assignment it accounts for \NCrowdComp{} general methods, \NCrowdCompAdopted{} inclusions and none in three or more frameworks. This comparison covers framework options only (Section~\ref{sec:limits}).

Most of the widely included methods are motivated by a failure that appears in long or large runs: entropy collapse, length bias, extreme importance ratios, instability in mixture-of-experts models, and the mismatch between rollout and training probabilities (Table~\ref{tab:wide}).

\section{Proposals and merges}
\label{sec:routes}

\begin{table}[!htbp]
\centering
\small
\setlength{\tabcolsep}{3pt}
\caption{\textbf{Methods and proposals by framework.} Unit: landing of a catalog method with a source from 2025 or later, and proposals for general methods that no framework included. Methods counts the landings in the framework, and first anywhere counts the methods with their first landing there. Median days run from the source to the landing and leave out upper bounds. A landing pull request counts as opened by a prior merger when its author had merged a pull request in the framework before opening it, and each pull request counts once. Days open run from opening to merge. Proposals are the coded issues and pull requests of Table~\ref{tab:attempts}, and lapsed proposals were closed as inactive or in a stale-issue cleanup. Declined proposals include those redirected elsewhere. ROLL publishes code in bulk releases without pull requests.}
\label{tab:frameworks}
\begin{tabular}{lrrrrrrrrr}
\toprule
 & & First & Median & \multicolumn{2}{c}{Landing pull requests} & \multicolumn{4}{c}{Proposals for other methods} \\
\cmidrule(lr){5-6} \cmidrule(lr){7-10}
Framework & Methods & anywhere & days & By a prior merger & Days open & All & Open & Declined & Lapsed \\
\midrule
verl & 30 & 21 & 25 & 1 of 27 & 10 & 20 & 12 & 0 & 7 \\
TRL & 24 & 16 & 35 & 13 of 23 & 5 & 18 & 3 & 10 & 2 \\
AReaL & 14 & 4 & 170 & 3 of 13 & 6 & 1 & 0 & 0 & 1 \\
NeMo-RL & 14 & 3 & 167 & 2 of 13 & 17 & 2 & 2 & 0 & 0 \\
ROLL & 12 & 4 & 99 & -- & -- & 0 & 0 & 0 & 0 \\
slime & 11 & 3 & 64 & 6 of 7 & 0 & 4 & 2 & 2 & 0 \\
OpenRLHF & 9 & 3 & 34 & 4 of 5 & 0 & 3 & 3 & 0 & 0 \\
\bottomrule
\end{tabular}

\end{table}

The features above describe the methods that got in. The public records of the frameworks show how the others fared, and the main finding is an absence: maintainers turned away few methods, \NRefusedPapers{} of \NNotIncluded{}, because few were ever proposed. We downloaded all \NThreads{} issues, pull requests and discussions of the seven frameworks on 18 October 2026 (UTC) and searched them for every general method in the pool. A paper has a record when a thread links its arXiv page, gives its arXiv identifier or quotes its full title.

Few methods appear in a framework's public record. Most general methods that no framework included have no record in any of the seven frameworks, \NNoRecord{} of \NNotIncluded{}, and another \NCitedOnly{} are cited only in passing, for example in lists of projects built on a framework. For \NImplPapers{} papers someone opened a pull request to implement the method, and for \NRequestPapers{} more someone requested it in an issue or a discussion. Chat groups and internal repositories leave no public trace, so a missing record does not show that maintainers never saw a paper.

Method names were checked separately, because many are ambiguous. Among the \NNoRecord{} general methods without a record, \NNameSearch{} have a method name that occurs once in the pool, and \NNameCand{} of these names appear in the threads. Reading those threads showed that \NNameRefers{} refer to the paper, \NNameProposal{} as proposals and \NNameCited{} in passing. The other matches share a name with something else, such as the NCCL SHARP plugin. Counting the \NNameRefers{} lowers the share without a record from \NNoRecordPct{} to \NNoRecordAfterPct{} percent, and applying the same rate to papers without a searchable name gives \NNoRecordExtrapPct{} percent. The coded proposals below come from link, identifier and title matches, so the \NAttemptPapers{} papers with a proposal are a lower bound: the name search found \NNameProposal{} more.

Many proposals stay open without a reply. We coded each of the \NAttemptThreads{} threads on these \NAttemptPapers{} papers from its own text (Table~\ref{tab:attempts}). By the snapshot, \NPROpen{} of the \NAttemptPRs{} pull requests were still open after a median of \OpenPRMedianAge{} days, and no prior merger had replied in \NPROpenNoReply{} of them. Maintainers declined \NPRDeclined{} pull requests and redirected \NPRRedirected{}, while the rest were withdrawn, closed as inactive or closed without a comment. Requests fared similarly: of the \NAttemptRequests{} in issues and discussions, a scheduled cleanup of inactive issues closed \NReqStale{}, and maintainers declined \NReqDeclined{} and redirected \NReqRedirected{}. In \NNoMergerReplyNone{} of the \NNoMergerReply{} open threads without a prior-merger reply, nobody except the proposer and automated reviewers had replied.

Merged pull requests drew concerns too. One reader coded the \RevMN{} merged landing pull requests and the \NAttemptThreads{} proposals with the same fields, each from the thread's own text (Table~\ref{tab:threads}). Someone other than the proposer raised a concern in \RevMAny{} merged pull requests, mostly about design or missing tests and documentation. The proposer revised the code in \RevMFixed{} of the \RevMN{}, and \RevMOpen{} still had an open concern when they were merged. Prior mergers raised the concerns in \RevMPrior{} of the \RevMAny{} merged threads with a concern and in \RevOtherPrior{} of the \RevOtherAny{} other threads with one.

\begin{table}[!htbp]
\centering
\small
\caption{\textbf{Review records of merged pull requests and of proposals that were not merged.} Unit: thread. Merged pull requests are the landing pull requests of recent methods, counted once each. Unmerged pull requests and requests are the proposals of Table~\ref{tab:attempts}. A concern is a problem that someone other than the proposer raises explicitly, and one thread can carry several.}
\label{tab:threads}
\begin{tabular}{lrrr}
\toprule
 & Merged & Unmerged & \\
Thread record & pull requests & pull requests & Requests \\
\midrule
Threads & 88 & 28 & 20 \\
\addlinespace
Reply from someone other than the proposer & 71 & 14 & 10 \\
Concern raised in the thread & 60 & 11 & 7 \\
\quad design or interface & 50 & 7 & 1 \\
\quad tests or documentation & 39 & 4 & 0 \\
\quad evidence & 18 & 3 & 1 \\
\quad placement & 8 & 4 & 4 \\
\quad maintenance burden & 6 & 5 & 6 \\
\quad thin demand & 0 & 3 & 5 \\
\quad already covered & 5 & 2 & 3 \\
Concern still open at the end & 41 & 9 & 7 \\
Training results shown & 26 & 9 & 2 \\
Results requested & 17 & 3 & 1 \\
Demand stated in the thread & 14 & 3 & 9 \\
\bottomrule
\end{tabular}

\end{table}

Concerns about demand and maintenance burden follow a different pattern: they appear in \RevMDS{} of \RevMN{} merged pull requests, against \RevUDS{} of \RevUN{} unmerged pull requests and \RevRDS{} of \RevRN{} requests. All of these unmerged pull requests come from TRL, so this pattern describes TRL more than the whole ecosystem. Training results appear about as often in merged and unmerged pull requests, at \RevMResults{} of \RevMN{} and \RevUResults{} of \RevUN{}. The largest difference is attention: \RevUSilent{} of the \RevUN{} unmerged pull requests drew no reply from anyone but the proposer, against \RevMSilent{} of the \RevMN{} merged ones. TRL weighs demand by the number of requests and reactions, and in one thread by whether other frameworks implement the method. slime keeps its main repository for changes that CI or routine training runs can verify, and it points other methods to separate repositories with reproducible recipes.

The frameworks handle proposals in different ways (Table~\ref{tab:frameworks}). In TRL, prior mergers opened \PrTRLMaint{} of the \PrTRLLand{} pull requests that brought methods in, and the project declined or redirected \PrTRLAttDeclined{} of \PrTRLAtt{} proposals for other methods. verl takes methods in from other contributors, often as recipes: prior mergers opened \PrverlMaint{} of its \PrverlLand{} landing pull requests, while \PrverlAttOpen{} of \PrverlAtt{} proposals for other methods remain open and \PrverlAttStale{} lapsed. slime and OpenRLHF add methods mainly through prior mergers, at \PrslimeMaint{} of \PrslimeLand{} and \PrOpenRLHFMaint{} of \PrOpenRLHFLand{} landing pull requests. NeMo-RL reviewers asked for training results in \RevNeMoMReq{} of its \RevNeMoMN{} merged pull requests.

Some pull requests brought in several methods, so the \NLandingEvents{} landings of recent methods with a pull request came from \NLandingPRs{} distinct pull requests. Paper authors made a commit in \NLandingAuthor{} of the landings, and \NLandingPriorRequest{} followed an earlier request in the same repository. Counting each pull request once, prior mergers opened \NPRPriorMerger{} of the \NLandingPRs{}. Merged pull requests had been open for a median of \PRDaysOpen{} days, a figure that covers only proposals that were merged. Inside each framework's own package they changed a median of \PRCoreLines{} lines, against \AttemptCoreLines{} for proposals that were not merged. The form of the change coded from the paper relates only weakly to this size: medians are \CoreLinesCoef{} lines for coefficient changes, \CoreLinesTerm{} for new terms and \CoreLinesPipe{} for pipeline changes, with each pull request counted once per form it carries.

Inclusion rarely reverses: \NRemoved{} of the \NRecentLandings{} landings of recent methods were later removed. TRL dropped the experimental trainers for GFPO, DPPO and PAPO, and AReaL dropped the AEnt recipe in a refactor. NeMo-RL reverted its first GSPO commit a day after it landed and added the method again with test fixes four days later.

One method arrived as a fix. In June 2025 a NeMo-RL developer implemented the KL gradient correction of \citet{arxiv250609477} behind a switch and withdrew it, because the experiments showed no gain. A year later a fix to the KL backward pass made the corrected gradient the behaviour of every run with a KL penalty, and its code cites the paper. We count this fix as the method's landing, as a conditional default.

\section{Failures that users report}
\label{sec:demand}

The frameworks' issues are where users would report the failures that the catalog's methods address, so they record what users name in public. We screened the \NIssueUniverse{} issues and discussions opened in the seven frameworks between January 2025 and the snapshot with keyword patterns for the thirteen failures of the catalog, and one reader coded the title and opening post of each of the \NIssueScreened{} matches (Appendix~\ref{app:record}). A thread counts when its opener reports observing the failure in a run. Most matches were bugs, usage questions or feature requests, and \NIssueFailure{} threads report a training failure (Table~\ref{tab:demand}). Among them, \NIssueInstability{} describe a collapse or divergence with no cause named, such as a loss that turns to NaN or a reward that drops to zero, and \NIssueTrainInferMismatch{} name the mismatch between rollout and training probabilities. A random sample of \NIssueRecall{} unscreened threads, read the same way, contained no failure report.

\begin{table}[!htbp]
\centering
\small
\caption{\textbf{Training failures in the frameworks' issues against the methods that address them.} Unit: thread in the first column, general method in the next two, catalog method in the last. Issue threads are issues and discussions opened since January 2025 with an opening post that reports the failure. General methods naming the failure are the pool's general methods with an abstract that names it, with the number of those that at least one framework includes. Catalog methods are those addressing the failure that at least one framework includes. The two rows with dashes are failures outside the catalog's list.}
\label{tab:demand}
\begin{tabular}{lrrrr}
\toprule
 & Issue threads & \multicolumn{2}{c}{General methods} & Catalog methods \\
\cmidrule(lr){3-4}
Failure & naming it & naming it & included & in a framework \\
\midrule
Collapse or divergence, cause not named & 23 & -- & -- & -- \\
Rollout-training mismatch & 12 & 20 & 3 & 8 \\
Response-length drift & 5 & 167 & 5 & 5 \\
Stale rollout data & 3 & 52 & 4 & 7 \\
KL penalty grows & 3 & -- & -- & -- \\
Groups without signal & 1 & 109 & 2 & 2 \\
MoE instability & 1 & 6 & 4 & 3 \\
Sparse reward & 0 & 218 & 4 & 3 \\
Loss of diversity & 0 & 156 & 2 & 3 \\
Gradient variance & 0 & 115 & 8 & 6 \\
Entropy collapse & 0 & 87 & 4 & 5 \\
Value-model cost & 0 & 31 & 6 & 6 \\
Extreme importance ratios & 0 & 18 & 5 & 7 \\
Multi-reward collapse & 0 & 14 & 1 & 1 \\
Truncation noise & 0 & 5 & 0 & 1 \\
\bottomrule
\end{tabular}

\end{table}

The failures named in public differ from those that papers address. The failures that abstracts address most often, sparse reward in \NSupplySparseReward{} general methods and the loss of diversity in \NSupplyDiversityLoss{}, appear in no screened thread. The mismatch between rollout and training probabilities is named by \NIssueTrainInferMismatch{} threads, and \NSupplyTrainInferMismatch{} general methods and \NCatalogTrainInferMismatch{} catalog methods in the frameworks address it, more than for any other failure. The record measures what users name in public and not what they need, because an abstract and an issue serve different purposes and the \NIssueInstability{} collapses with no cause named may belong to any row of the table. Requests for methods outnumber failure reports: \NIssueRequest{} threads ask for one, \NReqDynamicSampling{} of them for DAPO's dynamic sampling, \NReqMismatch{} for TIS or MIS and \NReqRoutingReplay{} for rollout routing replay, all methods that several frameworks already include. TRL weighs demand by such requests (Section~\ref{sec:routes}), and these counts show how thin that signal is: the most requested method drew \NReqDynamicSampling{} requests across seven frameworks in 21 months. The random sample of \NIssueRecall{} unscreened threads supports the screen and no more, so the counts are lower bounds, and a count of zero shows that nobody named the failure and not that nobody met it.

\section{Methods in large model reports}
\label{sec:use}

Inclusion describes framework code, so we read the technical reports of large open-weight models for the included methods their training used. The pool holds \NReports{} reports of open-weight models with 100B parameters or more, from \ReportFirst{} in January 2025 to \ReportLast{} in June 2026 (Table~\ref{tab:reports}). The pool requires reinforcement learning in the title or abstract, so reports that mention RL only in the body fall outside it: Qwen3 \citep{arxiv250509388}, the model behind GSPO, is one of them, and Appendix~\ref{app:record} lists the others we know of. For each report we read the RL sections and recorded the training algorithm, the framework it names and the widely included methods of Table~\ref{tab:wide} it states that it used. A method counts as stated when the report names it or cites its source, and a change described under another name counts as a variant.

\begin{table}[!htbp]
\centering
\footnotesize
\caption{\textbf{Widely included methods stated by technical reports of open-weight models with 100B parameters or more.} Unit: report in the pool. A check marks a method the report states it used, S marks the report that is the method's source, V a variant described under another name, and T a method tried and not kept. Framework stated is the training system the report names.}
\label{tab:reports}
\setlength{\tabcolsep}{2.5pt}
\begin{tabular}{@{}l l l ccccccccccccc@{}}
\toprule
Report & Size & Framework stated & \rotatebox{90}{Clip-higher} & \rotatebox{90}{GSPO} & \rotatebox{90}{Dyn. sampling} & \rotatebox{90}{Dr. GRPO} & \rotatebox{90}{CISPO} & \rotatebox{90}{TIS} & \rotatebox{90}{MIS} & \rotatebox{90}{R++} & \rotatebox{90}{Token loss} & \rotatebox{90}{Overlong} & \rotatebox{90}{R3} & \rotatebox{90}{SAPO} & \rotatebox{90}{GDPO} \\
\midrule
DeepSeek-R1~\citep{arxiv250112948} & 671B MoE & in-house &  &  &  &  &  &  &  &  &  &  &  &  &  \\
Llama-Nemotron~\citep{arxiv250500949} & 253B dense & NeMo-Aligner &  &  &  &  &  &  &  &  &  &  &  &  &  \\
MiniMax-M1~\citep{arxiv250613585} & 456B MoE & not stated & T &  & \checkmark &  & S &  &  &  & \checkmark & \checkmark &  &  &  \\
Kimi K2~\citep{arxiv250720534} & 1T MoE & in-house &  &  &  &  &  &  &  &  &  &  &  &  &  \\
gpt-oss~\citep{arxiv250810925} & 117B MoE & in-house &  &  &  &  &  &  &  &  &  &  &  &  &  \\
GLM-4.5~\citep{arxiv250806471} & 355B MoE & slime &  &  & \checkmark &  &  &  &  &  & \checkmark &  &  &  &  \\
LongCat-Flash-Thinking~\citep{arxiv250918883} & 560B MoE & DORA (in-house) & \checkmark &  & V & \checkmark &  &  &  &  & \checkmark & V &  &  &  \\
Ring-1T~\citep{arxiv251018855} & 1T MoE & AReaL &  &  &  &  &  & T & S &  &  &  &  &  &  \\
DeepSeek-V3.2~\citep{arxiv251202556} & 671B MoE & in-house &  &  &  &  &  &  &  &  &  &  & V &  &  \\
LongCat-Flash-Thinking-2601~\citep{arxiv260116725} & 560B MoE & DORA (in-house) &  & \checkmark &  &  &  &  &  &  &  &  &  &  &  \\
MiniMax-M2~\citep{arxiv260526494} & 230B MoE & in-house &  &  &  &  & \checkmark &  &  &  &  &  &  &  &  \\
Nemotron 3 Ultra~\citep{arxiv260615007} & 550B MoE & NeMo-RL &  &  &  &  &  &  & \checkmark &  &  &  &  &  &  \\
Ling and Ring 2.6~\citep{arxiv260615079} & 1T MoE & AReaL &  &  &  &  &  &  & V &  &  &  &  &  &  \\
\bottomrule
\end{tabular}

\end{table}

The reports name the included methods, but each names few. Of the \NReports{} reports, \NReportsUseAny{} state at least one of the \NWideMethods{} methods, \NReportsSource{} are the source of one, and \NReportsVariant{} describe a variant under their own name. DeepSeek's Keep Routing replays the rollout's expert routes as rollout routing replay does, LongCat filters all-correct and all-wrong prompts with replacement where DAPO samples without, and KPop replaces IcePop's ratio bounds with a binary KL test. Token-level loss aggregation is stated most often, by \NRepDapoTokenLoss{} reports. The KL term is removed in \NReportsNoKL{} reports, a change every framework allows through its KL coefficient and TRL made the default. None of the methods appears in \NReportsNone{} reports: DeepSeek-R1 and Llama-Nemotron predate most of them, Kimi K2 keeps the K1.5 objective, and gpt-oss gives no detail. No report states SAPO, GDPO or REINFORCE++, rollout routing replay appears only as a variant, and TIS appears only as a comparison. One of the seven frameworks is named in \NReportsNamedFramework{} reports: GLM-4.5 trains on slime, Ring-1T and Ling and Ring 2.6 on AReaL, and Nemotron 3 Ultra on NeMo-RL. Llama-Nemotron names NeMo-Aligner, the predecessor of NeMo-RL, and the other reports train on in-house systems. The reports are a second observable record of practice, and it overlaps with the frameworks' code: the stabilisers that six or seven frameworks ship are the ones the reports state. The overlap does not show the order of influence, because we read the latest version of each report, and the frameworks took in GSPO and MIS within weeks of the papers that introduced them.

\section{Implications for researchers and maintainers}
\label{sec:advice}

The comparison and the record describe past inclusion, and each paragraph below pairs a property of included methods with one step that a paper and its authors can take. None of these steps is shown to raise the chance of inclusion.

\paragraph{A public proposal.} Most methods never enter a framework's public record, while \NLandingPriorRequest{} of the \NLandingEvents{} landings followed an earlier issue or discussion that cited the paper. A record does not secure attention, since verl closed \PrverlAttStale{} of its \PrverlAtt{} proposals for other methods in scheduled cleanups. A short issue that names the training failure, the change and the evidence still gives maintainers something to find and answer.

\paragraph{Framework rules.} Each framework states its own rules. TRL asks for tests, documentation and an entry in its paper index, and it adds options to its main trainer only with visible demand. slime accepts changes that routine runs can verify, and verl collects many methods as recipes. General libraries can set stricter thresholds: the rule of thumb of scikit-learn for a new algorithm is three years since publication, 200 citations and wide use \citep{sklearnfaq}. The RL frameworks took methods in within a median of \LagMedian{} days, so their written rules and their review do the selecting. Merged pull requests drew concerns about design and tests in most cases, and their proposers often revised the code. A proposal that follows the written rules answers part of these concerns in advance.

\paragraph{Results that maintainers can rerun.} Training results in the thread did not separate merged from unmerged pull requests, but NeMo-RL reviewers asked for them in \RevNeMoMReq{} of \RevNeMoMN{} merged pull requests, and TRL asked for reproduced results before adding a method to its main trainer. A configuration, a seed and a short run on a public model let maintainers check that the method runs and shows its signal. Matching the paper's gains takes its baselines, data and budget.

\paragraph{The failure and its signal.} Included papers more often named an observed training failure, at \CCFailureCasePct{} against \CCFailureCtrlPct{} percent. Reporting the signal of that failure, such as the entropy curve or the divergence between rollout and training probabilities, lets others check the fix on their own runs.

\paragraph{Evidence at scale.} Included papers more often trained policies of 30B parameters or more. Teams without that compute can still target failures that larger runs report and publish the signal above, so that groups with more compute can test the fix.

\paragraph{Changes stated as configuration fields.} Included papers more often changed a coefficient, clip, mask or weight in the existing loss, and frameworks expose these methods as one or two configuration fields. A paper can name the configuration field it changes, the training stage it touches and any models it needs, so that maintainers can place the change without reading the code.

\paragraph{Choice of component.} Frameworks included \NCoreCompAdopted{} of \NCoreComp{} general methods on clipping, off-policy correction and loss aggregation, and \NCrowdCompAdopted{} of \NCrowdComp{} on rollout sampling and credit assignment. Methods in the second group can also live in user code, outside the options we track. In the frameworks' issues, the mismatch between rollout and training probabilities was named in \NIssueTrainInferMismatch{} failure reports and \NReqMismatch{} method requests, while no screened thread named sparse reward or the loss of diversity (Section~\ref{sec:demand}).

\paragraph{For maintainers.} The record also shows what a framework could change, under the same caveat. Proposals go unanswered more often than they are refused: \NPROpenNoReply{} of the \NPROpen{} open pull requests for other methods had no reply from a prior merger after a median of \OpenPRMedianAge{} days. Scheduled cleanups closed \NReqStale{} of \NAttemptRequests{} requests and \PrverlAttStale{} of \PrverlAtt{} proposals in verl. A stated route for methods, such as verl's recipes or slime's separate repositories, lets a maintainer answer a proposal with a destination instead of leaving it to the cleanup. A proposal template that asks for the failure, the change, the evidence and the configuration field would put in the first message the items that reviewers raised in \RevMAny{} of \RevMN{} merged pull requests. Demand measured by requests is thin (Section~\ref{sec:demand}), so a framework that waits for requests before adding an option will see few. The technical reports of large runs (Section~\ref{sec:use}) name the failures those runs met and the methods they used, a second public record a framework can read.

\section{Limitations}
\label{sec:limits}

Inclusion is not use: we observe implementations on main branches, and the data do not show the frequency of use or the results users get. Section~\ref{sec:use} reads what \NReports{} reports state, and a statement in a report is not code either. The numerator is a lower bound, because it counts only options we can parse and implementations that framework files cite. Sampling and reward logic can live outside framework options, so the component comparison covers framework options only. The pool is screened from three arXiv categories, and the seven frameworks represent neither closed training stacks nor every open one. Four other open frameworks meet the selection criteria, and their current code adds \NFwExtraNew{} general methods to the included set (Section~\ref{sec:inclusion}).

The comparison in Section~\ref{sec:what} is observational and small. Affiliations come from each paper's first page and match a framework at the level of the company, so a team that works with a framework's maintainers from another company counts as outside. The public record misses chat groups and internal repositories, and the search by method name reaches only names that are distinctive. ROLL and AReaL publish much of their code in bulk releases. Landings without a pull request still went through review, and the pull requests with a record do not represent each framework's whole process. Thread texts can be edited or deleted after the fact, and one reader coded the review fields of all threads. The failure reading of Section~\ref{sec:demand} covers titles and opening posts after a keyword screen, by one reader, so its counts are lower bounds, and a collapse reported without a cause may belong to any of the failures. The reports of Section~\ref{sec:use} cover \NReports{} models from nine organisations and state what their authors chose to write. Whether maintainers act on such reports, on requests or on their own runs is not measured here, so the overlap between the reports and the frameworks' code is an observation and not a direction of influence.

\section{The tracker}
\label{sec:tracker}

The data and code behind this report form a maintained tracker at \url{https://github.com/pth2002/rl-method-tracker}. Its index groups the catalog by the training failure each method addresses. For every method it shows the frameworks that include it, the tier and the day it landed, with the setting that selects the method and a link to the file and line at the analysed commit. A second page gives, for each framework, its contribution rules, the files that hold its algorithms, the pull requests that brought recent methods in, the concerns raised in their review and the outcome of earlier proposals. Readers can report a wrong record through a correction form in the repository. A weekly job pulls the frameworks, re-traces their history, verifies sources against arXiv and opens a pull request. An option the parser detects without a mapping fails validation, so a person reviews every addition before it reaches the index. The release fixes the samples, labels, full-text coding and framework commits behind this report, together with a checksum list. It also stores the record of proposals. For each general method it lists the issues and pull requests that cite it, their state at the snapshot and the coded outcome. Thread texts stay on GitHub, and each coded thread keeps one short quote as evidence for its code. The release also stores the coded failure reports of Section~\ref{sec:demand} and the readings of the model reports of Section~\ref{sec:use}, each with one quote. The weekly job also refreshes this record and lists new proposals for coding. A separate command checks the inputs against that list and reproduces the report's numbers, tables and figures without network access.

\section{Conclusion}

Most RL methods for language models reach practitioners as papers and not as options in the frameworks they train with. Between January 2025 and June 2026, \RateMain{} percent of the general RL training methods in our pool reached one of seven open-source frameworks, and three or more frameworks include \NThree{} of \NGeneral{} today. Across five quarterly cohorts the papers grew \QSupplyRatio{}-fold while those included within 180 days stayed at \QIncMin{} to \QIncMax{}. Included methods more often changed one part of the existing loss and more often came with experiments on large policies, while released code was as common among them as among the others. Merged pull requests drew concerns about design and tests as well, while concerns about demand and maintenance burden appeared mostly in proposals that were not merged. Yet few methods were turned away in public, because most never reached a framework's issues or pull requests at all. Users rarely report training failures in those issues, and the technical reports of large open-weight models state the methods that most frameworks include, so the public record of demand is thin and the record of practice at scale overlaps with the frameworks' code. Other fields with shared training or evaluation code can reuse this measurement, since it needs only paper versions, implementation histories and public proposals. The tracker lists where each framework keeps its algorithms, the concerns its reviewers raised and the outcome of earlier proposals, so that authors can bring a method that far.

\bibliographystyle{plainnat}
\bibliography{refs}

\appendix

\section{Labelling rules}
\label{app:rules}

Each paper receives exactly one label from its title and abstract.

\begin{table}[!htbp]
\centering
\small
\begin{tabularx}{\linewidth}{l >{\raggedright\arraybackslash}X}
\toprule
Label & Meaning \\
\midrule
RL method & A new or modified training-time RL component: loss or objective, advantage or baseline, clipping or trust region, importance weighting or off-policy correction, rollout sampling or filtering that feeds the gradient, task-agnostic reward shaping, KL or entropy regularisation, credit assignment, or an optimiser built for RL on language models. \\
RL application & Existing RL algorithms applied to a task or domain, including domain-specific rewards or verifiers. \\
Analysis & Empirical study, reproduction or ablation without a new method. \\
Systems & Training systems, asynchronous pipelines and throughput work with the objective unchanged. \\
Offline preference & Offline preference optimisation in the DPO family. \\
Distillation & The main objective matches a teacher. \\
Other & Surveys, benchmarks, inference-time methods, SFT objectives, generic reward-model fitting, and papers outside RL training of language models. \\
\bottomrule
\end{tabularx}
\end{table}

An RL method is \emph{general} when the component is the paper's main contribution and is presented for RL with language models at large. It is \emph{embedded} when it serves a paper about one task, domain or application. A component that works only with a domain-specific reward makes the paper an application. Two included sources served as examples in an early version of these rules. Counting them, method recognition is \RecallHit{} of \RecallAll{}.

\section{Manual review}
\label{app:review}

The review sample of \ReviewAll{} papers was drawn from the first labelled batches, stratified by label. Table~\ref{tab:review} compares review and labels.

\begin{table}[!htbp]
\centering
\small
\caption{Manual review against the labels.}
\label{tab:review}
\begin{tabular}{lrrr}
\toprule
Review \textbackslash{} Label & General & Embedded & Other \\
\midrule
General & 68 & 3 & 1 \\
Embedded & 6 & 12 & 32 \\
Other & 2 & 1 & 75 \\
\bottomrule
\end{tabular}
\end{table}

Within each label stratum, review found general methods in \ReviewGenHit{} of \ReviewGenN{} papers labelled general, \ReviewEmbHit{} of \ReviewEmbN{} labelled embedded, and \ReviewOthHit{} of \ReviewOthN{} labelled otherwise. Weighting these rates by the size of each stratum in the pool gives about \ReviewEstimate{} general methods. The main disagreement lies between embedded methods and applications, a boundary outside the denominator.

\section{Full-text coding}
\label{app:fulltext}

The comparison covers every included general method from 2025 and a random sample of general methods from 2025 that no framework included. Controls were drawn within each quarter, in proportion to the quarterly counts of included papers at sampling time. The quarterly counts are \CCQOneCase{}, \CCQTwoCase{}, \CCQThreeCase{} and \CCQFourCase{} for included papers and \CCQOneCtrl{}, \CCQTwoCtrl{}, \CCQThreeCtrl{} and \CCQFourCtrl{} for controls. Papers were coded in a shuffled order without access to inclusion status, and every field was read from the first arXiv version. The included papers found by the citation audit were coded later from their first versions. A second reader recoded all of them blind, together with \CCRecodedCtrls{} controls, and the comparison uses the blind codes. On the recoded controls the two readers agree on size thresholds, mixture of experts, extra models, compute and affiliation, and differ most on code release and failure motivation. With the recoded controls substituted, the two main gaps compare against \CCRecBigThirtyCtrl{} of \CCRecBigThirtyCtrlN{} and \CCRecCoefCtrl{} of \CCRecCoefCtrlN{} controls. arXiv withdrew the first version of \CCNoVOne{} control paper, and the comparison leaves it out. One coded change form was a placeholder for a prompt-only change and is treated as missing. Each paper was coded on the following fields.
\begin{itemize}
\item Affiliations on the first page, and whether the version links training code for the method.
\item Whether the method trained a policy of at least 14B and of at least 30B parameters, and whether it trained a mixture of experts.
\item The form of the change. A \emph{coefficient} change alters a coefficient, clip, mask or weight in the existing loss. A \emph{term} adds a loss term, estimator or sampling rule. A \emph{pipeline} change adds a training stage, a model or a system component.
\item The number of new hyperparameters, and whether the method needs an extra model.
\item Whether the paper is motivated by an observed training failure.
\item The evaluation domains, and whether compute is reported.
\end{itemize}
Fields a paper does not state were left empty and are excluded from the corresponding row of Figure~\ref{fig:casecontrol-all}. Papers without an empirical evaluation are excluded from the row on evaluation breadth.

\begin{table}[!htbp]
\centering
\small
\caption{\textbf{Sensitivity of the two largest gaps.} Unit: paper. Each cell gives included papers against non-included papers. The last two rows leave out included papers that the framework of their own company includes, and then every paper from a company that maintains one of the frameworks. $p$ is from a two-sided Fisher exact test on the two counts of the cell.}
\label{tab:sensitivity}
\setlength{\tabcolsep}{4pt}
\begin{tabular}{lllll}
\toprule
Analysis & Policy of 30B or more & $p$ & Coefficient change & $p$ \\
\midrule
Main & \CCBigThirtyCase/\CCBigThirtyCaseN{} vs \CCBigThirtyCtrl/\CCBigThirtyCtrlN & \CCPMainBigThirty & \CCCoefCase/\CCCoefCaseN{} vs \CCCoefCtrl/\CCCoefCtrlN  & \CCPMainCoef \\
Including the exceptions & \CCAllBigThirtyCase/\CCAllBigThirtyCaseN{} vs \CCBigThirtyCtrl/\CCBigThirtyCtrlN & \CCPAllBigThirty & \CCAllCoefCase/\CCAllCoefCaseN{} vs \CCCoefCtrl/\CCCoefCtrlN  & \CCPAllCoef \\
Without model reports & \CCNoRepBigThirtyCase/\CCNoRepBigThirtyCaseN{} vs \CCNoRepBigThirtyCtrl/\CCNoRepBigThirtyCtrlN & \CCPNoRepBigThirty & \CCNoRepCoefCase/\CCNoRepCoefCaseN{} vs \CCNoRepCoefCtrl/\CCNoRepCoefCtrlN  & \CCPNoRepCoef \\
With an industry author & \CCIndBigThirtyCase/\CCIndBigThirtyCaseN{} vs \CCIndBigThirtyCtrl/\CCIndBigThirtyCtrlN & \CCPIndBigThirty & \CCIndCoefCase/\CCIndCoefCaseN{} vs \CCIndCoefCtrl/\CCIndCoefCtrlN  & \CCPIndCoef \\
Missing values against gap & \CCBigThirtyCaseWorst/\CCCases{} vs \CCBigThirtyCtrlWorst/\CCControls & \CCPWorstBigThirty & \CCCoefCaseWorst/\CCCases{} vs \CCCoefCtrlWorst/\CCControls  & \CCPWorstCoef \\
Threshold of 14B & \CCBigFourteenCase/\CCBigFourteenCaseN{} vs \CCBigFourteenCtrl/\CCBigFourteenCtrlN & \CCPFourteenBigThirty & -- & \\
Before the citation audit & \CCOrigBigThirtyCase/\CCOrigBigThirtyCaseN{} vs \CCBigThirtyCtrl/\CCBigThirtyCtrlN & \CCPOrigBigThirty & \CCOrigCoefCase/\CCOrigCoefCaseN{} vs \CCCoefCtrl/\CCCoefCtrlN  & \CCPOrigCoef \\
Without own-company inclusion & \CCNoOwnBigThirtyCase/\CCNoOwnBigThirtyCaseN{} vs \CCBigThirtyCtrl/\CCBigThirtyCtrlN & \CCPNoOwnBigThirty & \CCNoOwnCoefCase/\CCNoOwnCoefCaseN{} vs \CCCoefCtrl/\CCCoefCtrlN  & \CCPNoOwnCoef \\
Without framework companies & \CCNoOrgBigThirtyCase/\CCNoOrgBigThirtyCaseN{} vs \CCNoOrgBigThirtyCtrl/\CCNoOrgBigThirtyCtrlN & \CCPNoOrgBigThirty & \CCNoOrgCoefCase/\CCNoOrgCoefCaseN{} vs \CCNoOrgCoefCtrl/\CCNoOrgCoefCtrlN  & \CCPNoOrgCoef \\
\bottomrule
\end{tabular}
\end{table}

\section{Included papers outside the general rule}
\label{app:exceptions}

These \NExceptions{} included papers do not meet the general rule from their abstracts, and the main rate leaves them out. Most are model reports or system papers that describe an included component only in the body. Keeping them out lets the denominator and the comparison of Section~\ref{sec:what} use one eligibility rule.

\begin{table}[!htbp]
\centering
\small
\caption{\textbf{Included papers that do not meet the general rule from their abstracts.} The attribution of masked importance sampling stays open among near-simultaneous sources.}
\label{tab:exceptions}
\begin{tabularx}{\linewidth}{l l l >{\raggedright\arraybackslash}X}
\toprule
Source & Abstract label & Model report & Included methods \\
\midrule
\citet{arxiv250507291} & RL method (embedded) & yes & Two-sided ratio clipping \\
\citet{arxiv250522312} & RL application & yes & Adaptive entropy control \\
\citet{arxiv250524298} & RL method (embedded) & no & Decoupled PPO objective \\
\citet{arxiv250613585} & RL method (embedded) & yes & CISPO \\
\citet{arxiv250805731} & RL method (embedded) & yes & AEPO \\
\citet{arxiv250808221} & Analysis & no & LitePPO \\
\citet{arxiv251018855} & RL method (embedded) & yes & Masked importance sampling (MIS) \\
\citet{arxiv251202556} & RL application & yes & Off-policy sequence masking, Importance-weighted unbiased KL gradient \\
\bottomrule
\end{tabularx}

\end{table}

\section{The public record}
\label{app:record}

The record covers \NThreads{} issues, pull requests and discussions downloaded through the GitHub API on 18 October 2026 (UTC), after the framework snapshots in Table~\ref{tab:candidates}. Every discussion was read in full, and none had more comments than one request returns. Formal review texts were read for the coded threads and the landing pull requests only. A thread matches a paper through a link to its arXiv, Hugging Face papers or alphaXiv page, its identifier after ``arXiv:'', its bare identifier, or its full title of at least five words. Identifiers are checked against the pool, so every match names a pool paper. The separate search by method name uses names of at least four characters with a capital letter that occur once in the pool, and every match for a paper without a record was read in its thread.

The citation audit reads every line added on each main branch and keeps the first commit that cites a paper in each file. All files citing a paper were checked by hand. Of the \NAuditUnmatched{} general methods cited in framework files but missed by parsing, \NAuditAdded{} sit next to an implementation, including the NeMo-RL fix to the KL gradient and ROLL's trajectory-level agent pipeline. Another \NAuditList{} appear in lists of projects or news, \NAuditReport{} in reference lists of technical report PDFs read as text, and \NAuditOther{} as a configuration note or a background reference.

Each proposal thread received one outcome and the reasons a maintainer stated explicitly in it, coded from the thread's own text. A first reader coded all threads, and a second reader coded them again blind to the first coding. The analysis uses the blind codes. The two codings agree on the outcome of \AgreeOutcome{} of \AgreeThreads{} threads as raw agreement, and most differences concern redirects and withdrawals. Outcomes follow a fixed order. A proposer who closes the thread has withdrawn it. A maintainer who sends the proposal to an example, an experimental folder or a separate repository has redirected it. An explicit refusal is a decline, and a closure by a scheduled rule is a stale cleanup. When several threads in one framework concern the same paper, such as a withdrawn first attempt and its replacement, we code the main thread. Demand covers the number of requests, reactions and implementations elsewhere. Maintenance burden covers new options, trainers or interfaces to keep. Placement covers a move to an experimental folder, a recipe, an example or a separate repository. Evidence covers results to reproduce or gains to show. The other reasons are missing tests or documentation, an objection to the interface or architecture of the change, and an existing feature that already covers the case. A stated reason is also a concern in the review fields of the same thread, and one adjudication resolved the two threads where the codings first disagreed. The review fields of Table~\ref{tab:threads} were coded for every merged landing pull request and every proposal with one set of rules. A concern counts when someone other than the proposer raises it explicitly, and requests about style or naming do not count. An open concern is one the thread leaves unaddressed at its end, including follow-up work named at the merge.

The failure reading of Section~\ref{sec:demand} screened every issue and discussion opened since 1 January 2025 with keyword patterns for the thirteen failures of the catalog. The patterns were applied to the title and the first 800 characters of the opening post after removing code blocks, logs and configuration lines. One reader coded each of the \NIssueScreened{} matches as a failure report with the failure named, a bug, a usage question, a feature request or other. A method request is a thread that asks for a named method. The recall sample draws \NIssueRecall{} unscreened threads at random with a fixed seed, and the release keeps each coded thread's number, title, keywords and code. The report reading of Section~\ref{sec:use} takes every model report in the pool for an open-weight model with 100B or more total parameters, reads its RL sections in the latest arXiv version, and keeps one quote per report with the methods it states. Reports that mention RL only in the body are outside the pool. Among the open-weight models with 100B or more parameters released in the window, the reports of Qwen3 \citep{arxiv250509388}, Pangu Ultra MoE \citep{arxiv250504519}, Intern-S1 \citep{arxiv250815763}, Command A \citep{arxiv250400698}, dots.llm1 \citep{arxiv250605767} and Step-3 \citep{arxiv250719427} fall outside it. This list comes from our own check of the models we know and not from a systematic search.

{\small
\begin{longtable}{@{}>{\raggedright\arraybackslash}p{2.2cm} l l l >{\raggedright\arraybackslash}p{2.5cm} >{\raggedright\arraybackslash}p{2.6cm}@{}}
\caption{\textbf{Proposals to include general methods that no framework included.} Unit: issue, pull request or discussion that proposes or requests the method, coded blind from its own text. The arXiv identifier tells apart methods with the same name. Reasons are those a maintainer stated in the thread.}\label{tab:attempts}\\
\toprule
Method & arXiv & Framework & Thread & Outcome & Stated reasons \\
\midrule
\endfirsthead
\toprule
Method & arXiv & Framework & Thread & Outcome & Stated reasons \\
\midrule
\endhead
VC-PPO decoupled GAE & 2503.01491 & verl & issue \#479 & open, under discussion & -- \\
AGPO & 2503.15952 & TRL & PR \#6368 & redirected & maintenance burden, placement, evidence, already covered \\
JEPO & 2503.19618 & TRL & PR \#5411 & declined & maintenance burden, demand \\
CPPO (completion pruning) & 2503.22342 & TRL & issue \#3291 & open, no prior-merger reply & -- \\
AdaRFT & 2504.05520 & verl & PR \#1021 & open, under discussion & -- \\
SSR & 2504.08837 & TRL & PR \#3496 & closed as inactive & -- \\
GRPO-LEAD & 2504.09696 & TRL & PR \#6222 & withdrawn by proposer & -- \\
Turn-level advantage & 2505.11821 & verl & issue \#1654 & closed in stale cleanup & -- \\
DisCO & 2505.12366 & verl & issue \#3157 & closed in stale cleanup & -- \\
DisCO & 2505.12366 & verl & PR \#3357 & open, under discussion & -- \\
KTAE & 2505.16826 & verl & issue \#1783 & closed in stale cleanup & -- \\
Intuitor & 2505.19590 & TRL & PR \#3507 & withdrawn by proposer & -- \\
VeriFree & 2505.21493 & TRL & issue \#3777 & declined & demand, maintenance burden \\
T-PPO & 2506.15050 & verl & PR \#3728 & open, no prior-merger reply & -- \\
RLPR & 2506.18254 & TRL & issue \#3928 & declined & demand, placement, maintenance burden \\
ESSA & 2507.04453 & verl & PR \#3637 & open, no prior-merger reply & -- \\
FR3E & 2507.07017 & OpenRLHF & issue \#1093 & open, no prior-merger reply & -- \\
FR3E & 2507.07017 & verl & issue \#2755 & closed in stale cleanup & -- \\
RLCR & 2507.16806 & TRL & issue \#3871 & declined & demand, maintenance burden, already covered \\
LightningRL & 2508.03680 & verl & issue \#3434 & closed in stale cleanup & -- \\
GTPO & 2508.03772 & TRL & issue \#5226 & redirected & demand, maintenance burden, placement, design, already covered \\
Pass@k Training & 2508.10751 & verl & issue \#3393 & closed in stale cleanup & -- \\
CHORD & 2508.11408 & slime & issue \#1843 & open, no prior-merger reply & -- \\
CHORD & 2508.11408 & verl & issue \#3223 & closed in stale cleanup & -- \\
HICRA & 2509.03646 & TRL & PR \#4997 & declined & demand, maintenance burden, design \\
PSPO & 2509.21282 & TRL & PR \#4548 & declined & demand, maintenance burden, tests and docs \\
Evolution Strategies & 2509.24372 & OpenRLHF & PR \#1214 & open, no prior-merger reply & -- \\
GDPO & 2510.08554 & NeMo-RL & PR \#3550 & open, no prior-merger reply & -- \\
SRL & 2510.25992 & TRL & issue \#4759 & declined & demand, maintenance burden, already covered \\
RLVE & 2511.07317 & slime & PR \#1020 & redirected & placement \\
HA-DW & 2601.08521 & TRL & PR \#4872 & closed as inactive & maintenance burden, placement, evidence, tests and docs \\
JustGRPO & 2601.15165 & NeMo-RL & PR \#4258 & open, no prior-merger reply & -- \\
CFPO & 2601.22801 & OpenRLHF & PR \#1184 & open, no prior-merger reply & -- \\
CFPO & 2601.22801 & TRL & PR \#5027 & declined & tests and docs \\
CFPO & 2601.22801 & verl & PR \#5252 & open, no prior-merger reply & -- \\
F-GRPO & 2602.06717 & slime & issue \#1565 & redirected & maintenance burden, placement, evidence \\
F-GRPO & 2602.06717 & verl & issue \#5256 & withdrawn by proposer & -- \\
F-GRPO & 2602.06717 & verl & PR \#5334 & open, no prior-merger reply & -- \\
Dr. MAS & 2602.08847 & AReaL & issue \#1114 & closed in stale cleanup & -- \\
HACPO & 2603.02604 & verl recipes & issue \#143 & open, no prior-merger reply & -- \\
FIPO & 2603.19835 & slime & PR \#1801 & open, no prior-merger reply & -- \\
TPO & 2604.06159 & TRL & PR \#5591 & closed without comment & -- \\
SNR-Aware Filtering & 2604.06268 & TRL & PR \#7242 & open, no prior-merger reply & -- \\
SNR-Aware Filtering & 2604.06268 & verl & PR \#7894 & open, under discussion & -- \\
POISE & 2605.07579 & verl recipes & PR \#157 & open, no prior-merger reply & -- \\
CoDaPO & 2606.07950 & verl & PR \#7497 & open, no prior-merger reply & -- \\
CPPO & 2606.10968 & verl & PR \#6731 & open, under discussion & -- \\
STARE & 2606.19236 & TRL & PR \#6167 & open, no prior-merger reply & -- \\
\bottomrule
\end{longtable}

}

\section{Framework selection}
\label{app:frameworks}

Table~\ref{tab:candidates} applies the selection criteria of Section~\ref{sec:inclusion} to the open RL frameworks with at least \MinStars{} GitHub stars and gives the commit analysed in each. The recipe repository of verl is analysed at commit \VerlRecipeCommit{} of \VerlRecipeDate{}.

\begin{table}[H]
\centering
\small
\caption{\textbf{Open RL frameworks against the selection criteria of Section~\ref{sec:inclusion}.} Unit: repository with at least \MinStars{} stars on 19 October 2026. Commits are the snapshots of the seven tracked frameworks and the latest commits of the other qualifying ones. For qualifying frameworks outside the study, the count gives general methods that this commit implements and none of the seven include.}
\label{tab:candidates}
\begin{tabularx}{\linewidth}{@{}l r l >{\raggedright\arraybackslash}X@{}}
\toprule
Repository & Stars & Commit & Status \\
\midrule
\texttt{verl-project/verl} & 23,926 & \texttt{18cfb74852} 2026-10-16 & tracked \\
\texttt{huggingface/trl} & 19,551 & \texttt{6d6901c5e3} 2026-10-16 & tracked \\
\texttt{OpenRLHF/OpenRLHF} & 10,097 & \texttt{dc2a7ad326} 2026-09-17 & tracked \\
\texttt{THUDM/slime} & 8,686 & \texttt{859938dcb2} 2026-10-17 & tracked \\
\texttt{areal-project/AReaL} & 5,843 & \texttt{bb13e3bdb9} 2026-10-16 & tracked \\
\texttt{alibaba/ROLL} & 3,430 & \texttt{d96d0cce56} 2026-10-13 & tracked \\
\texttt{NVIDIA-NeMo/RL} & 2,074 & \texttt{83385307ab} 2026-10-17 & tracked \\
\texttt{RLinf/RLinf} & 5,439 & \texttt{a78b6d7f64} 2026-10-17 & meets the criteria; adds no general method \\
\texttt{allenai/open-instruct} & 3,901 & \texttt{ff32794a34} 2026-10-17 & meets the criteria; adds 1 general method \\
\texttt{NovaSky-AI/SkyRL} & 2,434 & \texttt{2c54cefc7c} 2026-10-16 & meets the criteria; adds no general method \\
\texttt{PrimeIntellect-ai/prime-rl} & 2,175 & \texttt{8dede158a9} 2026-10-19 & meets the criteria; adds 1 general method \\
\texttt{unslothai/unsloth} & 78,250 &  & general fine-tuning; its RL trainers build on TRL \\
\texttt{hiyouga/LlamaFactory} & 75,583 &  & general fine-tuning \\
\texttt{huggingface/open-r1} & 26,475 &  & a recipe that trains with TRL \\
\texttt{modelscope/ms-swift} & 15,886 &  & general fine-tuning; its RL trainers build on TRL \\
\texttt{OpenPipe/ART} & 10,813 &  & trains through Unsloth and TRL \\
\texttt{rllm-org/rllm} & 5,887 &  & trains through verl \\
\texttt{meta-pytorch/torchtune} & 5,818 &  & general fine-tuning \\
\texttt{PrimeIntellect-ai/verifiers} & 4,717 &  & an environment library \\
\texttt{radixark/miles} & 3,165 &  & a fork of slime \\
\bottomrule
\end{tabularx}

\end{table}

\section{Supplementary tables and figures}

\begin{table}[H]
\centering
\small
\caption{\textbf{Option names at the snapshot for recent methods included in three or more frameworks.} Unit: catalog method. Names are the registry keys, loss types, configuration fields or recipe folders that the parser maps to each method.}
\label{tab:configs}
\begin{tabularx}{\linewidth}{>{\raggedright\arraybackslash}p{3.2cm} >{\raggedright\arraybackslash}X}
\toprule
Method & Option names at the snapshot \\
\midrule
Clip-higher & verl: \texttt{clip\_ratio\_high}, \texttt{recipe/dapo}; TRL: \texttt{epsilon\_high}; OpenRLHF: \texttt{eps\_clip\_low\_high}; NeMo-RL: \texttt{ratio\_clip\_max}; ROLL: \texttt{pg\_clip\_high}; slime: \texttt{eps\_clip\_high}; AReaL: \texttt{dapo\_dynamic\_bs}, \texttt{eps\_clip\_higher} \\ \addlinespace[3pt]
GSPO & verl: \texttt{gspo}; TRL: \texttt{importance\_sampling\_level}; OpenRLHF: \texttt{gspo}; NeMo-RL: \texttt{sequence\_level\_importance\_ratios}; ROLL: \texttt{importance\_sampling\_seq}; slime: \texttt{gspo}; AReaL: \texttt{gspo}, \texttt{importance\_sampling\_level} \\ \addlinespace[3pt]
Dynamic sampling & verl: \texttt{filter\_groups}, \texttt{recipe/dapo}; OpenRLHF: \texttt{dynamic\_filtering}; NeMo-RL: \texttt{use\_dynamic\_sampling}; ROLL: \texttt{difficulty\_mask}; slime: \texttt{dynamic\_sampling\_filter}; AReaL: \texttt{dapo\_dynamic\_bs}, \texttt{dynamic\_bs} \\ \addlinespace[3pt]
Dr. GRPO & verl: \texttt{norm\_adv\_by\_std\_in\_grpo}, \texttt{seq-mean-token-sum-norm}; TRL: \texttt{dr\_grpo}, \texttt{scale\_rewards}; OpenRLHF: \texttt{dr\_grpo}, \texttt{no\_std\_norm}; ROLL: \texttt{seq-mean-token-sum-norm}; slime: \texttt{disable\_grpo\_std\_normalization}; AReaL: \texttt{drgrpo} \\ \addlinespace[3pt]
CISPO & verl: \texttt{cispo}; TRL: \texttt{cispo}; NeMo-RL: \texttt{use\_cispo}; ROLL: \texttt{cispo}; slime: \texttt{cispo}; AReaL: \texttt{use\_cispo\_loss} \\ \addlinespace[3pt]
Truncated importance sampling (TIS) & verl: \texttt{rollout\_is}; TRL: \texttt{vllm\_importance\_sampling\_correction}; OpenRLHF: \texttt{is\_correction}; NeMo-RL: \texttt{truncated\_importance\_sampling\_ratio}; ROLL: \texttt{tis}, \texttt{train\_infer\_correction}; slime: \texttt{use\_tis} \\ \addlinespace[3pt]
Masked importance sampling (MIS) & verl: \texttt{rollout\_rs}; TRL: \texttt{vllm\_importance\_sampling\_mode}; OpenRLHF: \texttt{is\_correction\_mask}; NeMo-RL: \texttt{seq\_logprob\_error\_threshold}, \texttt{seq\_mask\_tis}; ROLL: \texttt{train\_infer\_filter}; AReaL: \texttt{icepop}, \texttt{rejection\_sampling} \\ \addlinespace[3pt]
REINFORCE++ & verl: \texttt{reinforce\_plus\_plus}, \texttt{reinforce\_plus\_plus\_baseline}; OpenRLHF: \texttt{reinforce}, \texttt{reinforce\_baseline}; NeMo-RL: \texttt{reinforce\_plus\_plus}; slime: \texttt{reinforce\_plus\_plus}, \texttt{reinforce\_plus\_plus\_baseline}; AReaL: \texttt{reinforce\_baseline} \\ \addlinespace[3pt]
Token-level loss aggregation & verl: \texttt{recipe/dapo}; TRL: \texttt{dapo}; NeMo-RL: \texttt{token\_level\_loss}; ROLL: \texttt{token-mean}; slime: \texttt{calculate\_per\_token\_loss} \\ \addlinespace[3pt]
Overlong filtering and penalty & verl: \texttt{overlong\_buffer}, \texttt{recipe/dapo}; TRL: \texttt{mask\_truncated\_completions}; OpenRLHF: \texttt{overlong\_buffer}; NeMo-RL: \texttt{overlong\_filtering}; AReaL: \texttt{overlong\_reward\_penalty} \\ \addlinespace[3pt]
Rollout routing replay & verl: \texttt{enable\_rollout\_routing\_replay}; NeMo-RL: \texttt{router\_replay}; ROLL: \texttt{enable\_rollout\_routing\_replay}; slime: \texttt{use\_rollout\_routing\_replay}; AReaL: \texttt{enable\_router\_replay} \\ \addlinespace[3pt]
SAPO & verl: \texttt{sapo}; TRL: \texttt{sapo}; AReaL: \texttt{sapo}, \texttt{use\_sapo\_loss} \\ \addlinespace[3pt]
GDPO & verl: \texttt{gdpo}; TRL: \texttt{multi\_objective\_aggregation}; NeMo-RL: \texttt{gdpo} \\ \addlinespace[3pt]
\bottomrule
\end{tabularx}

\end{table}

\begin{figure}[!htbp]
\centering
\includegraphics[width=\linewidth]{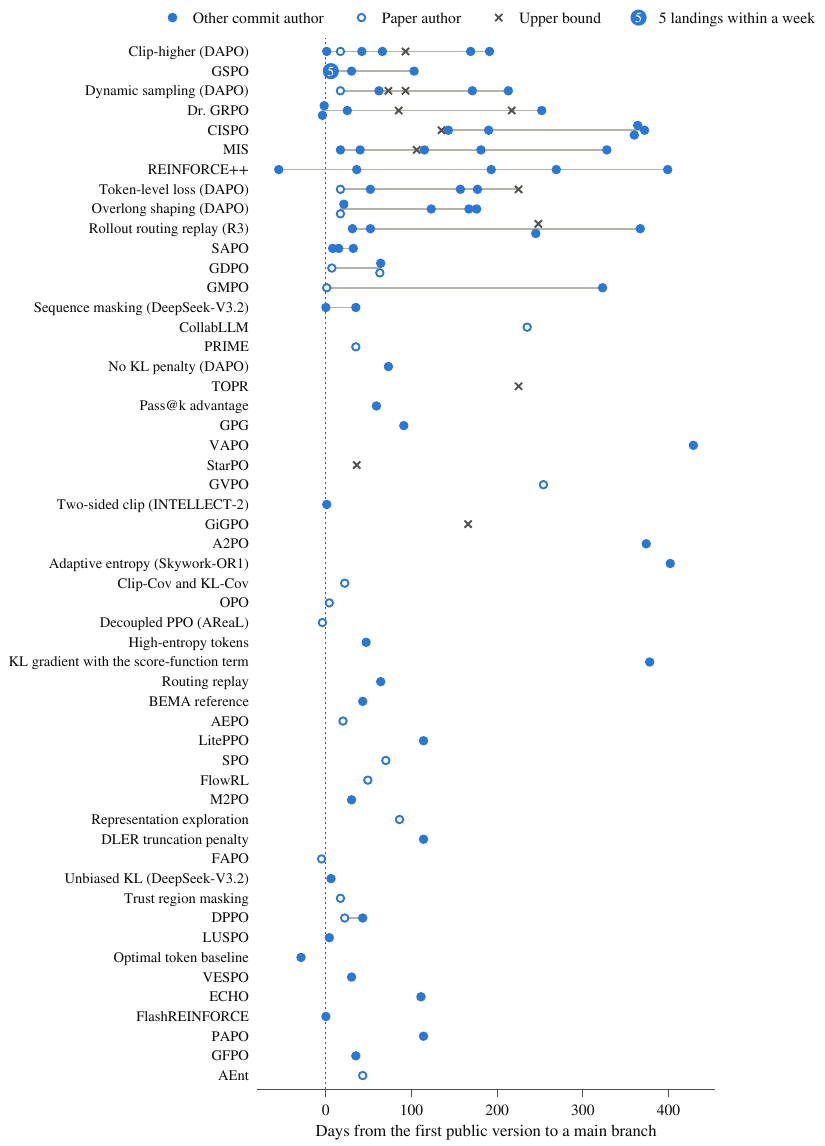}
\caption{\textbf{Time from first public version to a main branch, all methods.} Unit: catalog method with a dated source from 2025 or later. Marks as in Figure~\ref{fig:landing}.}
\label{fig:landing-all}
\end{figure}

\clearpage
\section{Method catalog}

{\small
\begin{longtable}{@{}>{\raggedright\arraybackslash}p{3.5cm} >{\raggedright\arraybackslash}p{2.3cm} l l >{\raggedright\arraybackslash}p{2.9cm}@{}}
\caption{\textbf{Method catalog.} Unit: catalog method. First landing is the earliest main-branch commit in any framework, with $\le$ for upper bounds. Frameworks list where the method is offered at the snapshot.}\label{tab:catalog} \\
\toprule
Method & Source & First public & First landing & Frameworks now \\
\midrule
\endfirsthead
\toprule
Method & Source & First public & First landing & Frameworks now \\
\midrule
\endhead
\bottomrule
\endfoot
PPO (clipped surrogate, GAE) & \citet{arxiv170706347} & 2017-07-20 & 2022-12-30 & verl, OpenRLHF, NeMo-RL, ROLL, slime, AReaL \\
Dual-clip PPO & \citet{arxiv191209729} & 2019-12-20 & 2025-04-02 & verl, OpenRLHF, NeMo-RL, ROLL, slime, AReaL \\
k3 (low-variance) KL estimator & \citet{schulman2020kl} & 2020-03-07 & 2025-01-23 & verl \\
ReMax & \citet{arxiv231010505} & 2023-10-16 & 2025-02-10 & verl \\
GRPO & \citet{arxiv240203300} & 2024-02-05 & 2025-01-20 & verl, TRL, OpenRLHF, NeMo-RL, ROLL, slime, AReaL \\
RLOO & \citet{arxiv240214740} & 2024-02-22 & 2024-05-22 & verl, TRL, OpenRLHF, NeMo-RL, AReaL \\
Direct Reward Optimisation (quadratic log-ratio penalty) & \citet{arxiv240519107} & 2024-05-29 & 2026-08-04 & verl \\
REINFORCE++ & \citet{arxiv250103262} & 2025-01-04 & 2024-11-10 & verl, OpenRLHF, NeMo-RL, slime, AReaL \\
CollabLLM (multiturn-aware rewards) & \citet{arxiv250200640} & 2025-02-02 & 2025-09-25 & verl \\
PRIME (implicit process rewards) & \citet{arxiv250201456} & 2025-02-03 & 2025-03-10 & verl \\
Clip-higher (decoupled clip range) & \citet{arxiv250314476} & 2025-03-18 & 2025-03-19 & verl, TRL, OpenRLHF, NeMo-RL, ROLL, slime, AReaL \\
Dynamic sampling (filter zero-variance groups) & \citet{arxiv250314476} & 2025-03-18 & 2025-04-04 & verl, OpenRLHF, NeMo-RL, ROLL, slime, AReaL \\
Remove the KL penalty for reasoning RL & \citet{arxiv250314476} & 2025-03-18 & 2025-05-30 & TRL \\
Overlong filtering and penalty & \citet{arxiv250314476} & 2025-03-18 & 2025-04-04 & verl, TRL, OpenRLHF, NeMo-RL, AReaL \\
Token-level loss aggregation & \citet{arxiv250314476} & 2025-03-18 & 2025-04-04 & verl, TRL, NeMo-RL, ROLL, slime \\
TOPR (tapered off-policy REINFORCE) & \citet{arxiv250314286} & 2025-03-18 & $\le$2025-10-29 & ROLL \\
Pass@k advantage & \citet{arxiv250319595} & 2025-03-25 & 2025-05-23 & verl \\
Dr. GRPO (no length or std normalisation) & \citet{arxiv250320783} & 2025-03-26 & 2025-03-22 & verl, TRL, OpenRLHF, ROLL, slime, AReaL \\
GPG & \citet{arxiv250402546} & 2025-04-03 & 2025-07-03 & verl \\
VAPO (decoupled and length-adaptive GAE, positive-example NLL) & \citet{arxiv250405118} & 2025-04-07 & 2026-06-10 & NeMo-RL \\
StarPO (trajectory-level optimisation for multi-turn agents, RAGEN) & \citet{arxiv250420073} & 2025-04-24 & $\le$2025-05-30 & ROLL \\
GVPO (group variance policy optimisation) & \citet{arxiv250419599} & 2025-04-28 & 2026-01-07 & verl \\
Two-sided ratio clipping (INTELLECT-2) & \citet{arxiv250507291} & 2025-05-12 & 2025-05-13 & TRL \\
GiGPO (group-in-group credit for agents) & \citet{arxiv250510978} & 2025-05-16 & $\le$2025-10-29 & ROLL \\
A2PO (optimal advantage regression) & \citet{arxiv250520686} & 2025-05-27 & 2026-06-05 & TRL \\
Clip-Cov and KL-Cov & \citet{arxiv250522617} & 2025-05-28 & 2025-06-19 & verl \\
Adaptive entropy control (Skywork-OR1) & \citet{arxiv250522312} & 2025-05-28 & 2026-07-04 & TRL \\
OPO (optimal reward baseline) & \citet{arxiv250523585} & 2025-05-29 & 2025-06-02 & verl \\
Decoupled PPO objective (AReaL) & \citet{arxiv250524298} & 2025-05-30 & 2025-05-26 & AReaL \\
Train on high-entropy tokens only & \citet{arxiv250601939} & 2025-06-02 & 2025-07-19 & TRL \\
KL gradient with the score-function term (Tang and Munos) & \citet{arxiv250609477} & 2025-06-11 & 2026-06-24 & NeMo-RL \\
CISPO & \citet{arxiv250613585} & 2025-06-16 & $\le$2025-10-29 & verl, TRL, NeMo-RL, ROLL, slime, AReaL \\
PAPO (implicit perception loss) & \citet{arxiv250706448} & 2025-07-08 & 2025-10-30 & removed \\
GSPO (sequence-level ratio) & \citet{arxiv250718071} & 2025-07-24 & 2025-07-27 & verl, TRL, OpenRLHF, NeMo-RL, ROLL, slime, AReaL \\
Routing replay across updates & \citet{arxiv250718071} & 2025-07-24 & 2025-09-26 & slime \\
GMPO (geometric-mean objective) & \citet{arxiv250720673} & 2025-07-28 & 2025-07-29 & verl, TRL \\
Bias-corrected EMA reference model & \citet{arxiv250800180} & 2025-07-31 & 2025-09-12 & TRL \\
Truncated importance sampling (TIS) & \citet{yao2025tis} & 2025-08 & 2025-08-06 & verl, TRL, OpenRLHF, NeMo-RL, ROLL, slime \\
AEPO (adaptive exploration policy optimisation) & \citet{arxiv250805731} & 2025-08-07 & 2025-08-27 & verl \\
LitePPO (group-mean, batch-std normalisation with token-level loss) & \citet{arxiv250808221} & 2025-08-11 & 2025-12-03 & AReaL \\
GFPO (group filtered policy optimisation) & \citet{arxiv250809726} & 2025-08-13 & 2025-09-17 & removed \\
AEnt (clamped entropy bonus with adaptive coefficient) & \citet{arxiv250903493} & 2025-09-03 & 2025-10-16 & removed \\
SPO (single-stream policy optimisation) & \citet{arxiv250913232} & 2025-09-16 & 2025-11-25 & verl \\
FlowRL (reward distribution matching) & \citet{arxiv250915207} & 2025-09-18 & 2025-11-06 & verl \\
M2PO (second-moment trust region for stale data) & \citet{arxiv251001161} & 2025-10-01 & 2025-10-31 & AReaL \\
Rollout routing replay (R3) & \citet{arxiv251011370} & 2025-10-13 & 2025-11-13 & verl, NeMo-RL, ROLL, slime, AReaL \\
Representation-based exploration & \citet{arxiv251011686} & 2025-10-13 & 2026-01-07 & verl \\
DLER truncation penalty & \citet{arxiv251015110} & 2025-10-16 & 2026-02-07 & NeMo-RL \\
Masked importance sampling (MIS) & \citet{arxiv251018855} & 2025-10-21 & 2025-11-07 & verl, TRL, OpenRLHF, NeMo-RL, ROLL, AReaL \\
FAPO & \citet{arxiv251022543} & 2025-10-26 & 2025-10-21 & verl \\
SAPO (soft adaptive gating) & \citet{arxiv251120347} & 2025-11-25 & 2025-12-03 & verl, TRL, AReaL \\
Off-policy sequence masking (DeepSeek-V3.2) & \citet{arxiv251202556} & 2025-12-02 & 2025-12-02 & TRL, slime \\
Importance-weighted unbiased KL gradient (DeepSeek-V3.2) & \citet{arxiv251202556} & 2025-12-02 & 2025-12-08 & TRL \\
Trust region masking (sequence max divergence) & \citet{arxiv251223075} & 2025-12-28 & 2026-01-14 & verl \\
GDPO (decoupled per-reward normalisation) & \citet{arxiv260105242} & 2026-01-08 & 2026-01-15 & verl, TRL, NeMo-RL \\
DPPO (divergence-based trust region) & \citet{arxiv260204879} & 2026-02-04 & 2026-02-26 & verl \\
LUSPO (length-unbiased sequence policy optimisation) & \citet{arxiv260205261} & 2026-02-05 & 2026-02-09 & TRL \\
Optimal token baseline with variance proxy & \citet{arxiv260207078} & 2026-02-06 & 2026-01-08 & verl \\
VESPO & \citet{arxiv260210693} & 2026-02-11 & 2026-03-13 & TRL \\
ECHO (auxiliary world-model loss on observation tokens) & \citet{arxiv260524517} & 2026-05-23 & 2026-09-11 & verl \\
FlashREINFORCE (single-rollout async RL) & framework & 2026-09-14 & 2026-09-14 & OpenRLHF \\
GRPO with replay buffer & framework & -- & 2025-09-24 & removed \\
KPop (binary-KL token masking for rollout-training mismatch) & framework & -- & 2026-06-15 & AReaL \\
Untruncated importance weights between rollout and trainer log-probs & framework & -- & 2025-04-22 & NeMo-RL \\
SAO (direct double-sided importance sampling, ROLL) & framework & -- & $\le$2026-09-29 & ROLL \\
\end{longtable}

}

\end{document}